\pdfoutput=1
\PassOptionsToPackage{draft}{minted}
\documentclass{arxivtmpl}

\let\templatebibliographystyle\bibliographystyle
\renewcommand{\bibliographystyle}[1]{}

\def\1{\bm{1}}

\DeclareMathAlphabet{\mathsfit}{\encodingdefault}{\sfdefault}{m}{sl}
\SetMathAlphabet{\mathsfit}{bold}{\encodingdefault}{\sfdefault}{bx}{n}

\tcbuselibrary{skins,breakable}
\newtcolorbox{findingbox}{enhanced,breakable,frame hidden,colback=gred!6,borderline west={3pt}{0pt}{gred},sharp corners,boxsep=0pt,left=12pt,right=10pt,top=8pt,bottom=8pt,before skip=12pt,after skip=12pt}
\newtcolorbox{notebox}{enhanced,breakable,frame hidden,colback=gblue!6,borderline west={3pt}{0pt}{gblue},sharp corners,boxsep=0pt,left=12pt,right=10pt,top=8pt,bottom=8pt,before skip=12pt,after skip=12pt}
\newtcolorbox{tipbox}{enhanced,breakable,frame hidden,colback=ggreen!6,borderline west={3pt}{0pt}{ggreen},sharp corners,boxsep=0pt,left=12pt,right=10pt,top=8pt,bottom=8pt,before skip=12pt,after skip=12pt}
\newtcolorbox{warnbox}{enhanced,breakable,frame hidden,colback=gyellow!6,borderline west={3pt}{0pt}{gyellow},sharp corners,boxsep=0pt,left=12pt,right=10pt,top=8pt,bottom=8pt,before skip=12pt,after skip=12pt}

\tcbuselibrary{listings}
\newcommand{\boxfont}{\fontsize{9}{11}\selectfont}
\newtcblisting{promptbox}[1][]{%
  enhanced, listing only, breakable, frame hidden, width=\linewidth,
  colback=gblue!6, borderline west={3pt}{0pt}{gblue},
  sharp corners, boxsep=0pt,
  left=12pt, right=10pt, top=8pt, bottom=8pt, before skip=12pt, after skip=12pt,
  before upper={\def\promptlabel{#1}\ifx\promptlabel\empty\else{\sffamily\bfseries\color{gblue}#1}\par\smallskip\fi},   
  listing options={basicstyle=\ttfamily\boxfont, escapeinside={(*}{*)}},
}

\newtcolorbox{transcriptbox}{%
  enhanced, breakable, frame hidden,
  colback=ggreen!6, borderline west={3pt}{0pt}{ggreen},
  sharp corners, boxsep=0pt, fontupper=\ttfamily\boxfont, halign upper=flush left,   
  left=12pt, right=10pt, top=6pt, bottom=8pt, before skip=12pt, after skip=12pt,
}

\colorlet{codebg}{gyellow!6}
\setminted{fontsize=\boxfont, breaklines, autogobble, tabsize=4, style=vs}
\newminted[codebox]{python}{bgcolor=codebg, frame=leftline, framerule=2.5pt, rulecolor=gyellow, framesep=2.5mm}

\hypersetup{
    colorlinks=true,
    linkcolor=linkcol,
    citecolor=linkcol,
    urlcolor=linkcol,
    pdfborderstyle={},
    pdfborder={0 0 0}
}
\renewcommand\Affilfont{\rmfamily\mdseries\fontsize{11.5}{14}\selectfont}

\colorlet{perfhue}{gblue}
\colorlet{perf0}{white}
\colorlet{perf10}{perfhue!12}
\colorlet{perf20}{perfhue!22}
\colorlet{perf30}{perfhue!33}
\colorlet{perf40}{perfhue!44}
\colorlet{perf50}{perfhue!55}
\colorlet{perf60}{perfhue!66}
\colorlet{perf70}{perfhue!77}
\colorlet{perf80}{perfhue!88}

\newcommand{\RN}[1]{%
	\textup{\lowercase\expandafter{\it \romannumeral#1}}%
}

\SetKwInput{KwInput}{Input}
\SetKwInput{KwOutput}{Output}
\DontPrintSemicolon

\SetKwComment{Comment}{\color{ggreen}\# }{}

\SetKwProg{Function}{def}{:}{}

\SetKwProg{For}{for}{:}{}
\SetKwProg{If}{if}{:}{}

\let\bibliographystyle\templatebibliographystyle
\usepackage[T1]{fontenc}

\usepackage{subcaption}
\usepackage{booktabs}
\usepackage{adjustbox}
\usepackage{tabularx}

\usepackage{tikz}
\usetikzlibrary{arrows.meta}

\usepackage{tcolorbox}
\tcbset{
    fmMode/.style={
    colframe=black,
    boxrule=1pt,
    arc=2pt,
    left=2pt, right=2pt,
    top=2pt, bottom=2pt,
    after skip=0pt, before skip=0pt:},
    noPadding/.style={
        boxsep=0pt, left=0pt, right=0pt, top=0pt, bottom=0pt
    }
}

\usepackage{listings}
\usepackage{url}
\usepackage{xspace}
\usepackage{xcolor}
\usepackage{float}
\usepackage{multirow}
\usepackage{enumitem}
\usepackage{makecell}
\usepackage{wrapfig}

\graphicspath{{images/}}

\newcommand{\geobench}{GeoGuesser Bench\xspace}

\newcommand{\typeCompetition}{Competition\xspace}
\newcommand{\numCompetition}{42\xspace}
\newcommand{\humanScore}{4770.3\xspace}

\newcommand{\typeOne}{Precise Localization\xspace}
\newcommand{\typeTwo}{Place Naming\xspace}
\newcommand{\typeThree}{Navigation Challenge\xspace}
\newcommand{\taskNameCell}[2][22mm]{\parbox[c]{#1}{\centering #2}}
\newcommand{\se}[1]{\,{\normalfont\scriptsize$\pm$#1}}
\newcommand{\numTypeOne}{80\xspace}
\newcommand{\numTypeTwo}{80\xspace}
\newcommand{\numTypeThree}{80\xspace}

\newcommand{\findingTypeThree}{Identification\xspace}
\newcommand{\countingTypeThree}{Enumeration\xspace}

\newcommand*{\codedatanotetext}{}

\renewcommand{\footnoterule}{\kern-3pt\hrule width\columnwidth\kern2.6pt}

\titlespacing*{\section}{0pt}{1.1\baselineskip}{0.35\baselineskip}
\titlespacing*{\subsection}{0pt}{0.85\baselineskip}{0.15\baselineskip}
\titleformat{\section}{\rmfamily\bfseries\Large}{\thesection}{0.7em}{}
\titleformat{\subsection}{\rmfamily\bfseries\large}{\thesubsection}{0.7em}{}
\setlist[itemize]{topsep=4pt,itemsep=3pt,parsep=0pt}

\fancypagestyle{fancy}{%
  \fancyhf{}%
  \fancyhead[C]{\fontsize{9}{11}\selectfont\color{black!50}%
    GeoGuesser Bench: Benchmarking Vision--Language Agents on Interactive Geolocation}%
  \fancyfoot[C]{\small\thepage}%
  \renewcommand{\footrulewidth}{0pt}}
\makeatletter
\fancypagestyle{firstpagestyle}{%
  \fancyhf{}%
  \fancyfoot[L]{\codedatanotetext}%
  \renewcommand{\footrulewidth}{\ifx\codedatanotetext\@empty 0pt\else 0.3pt\fi}}
\makeatother

\renewcommand{\titlefont}{\normalfont\bfseries\fontsize{22}{26}\selectfont}

\renewcommand\Affilfont{\rmfamily\mdseries\fontsize{10.5}{13}\selectfont}

\makeatletter
\renewcommand\AB@authnote[1]{\expandafter\ifblank\expandafter{#1}{}{\,\textsuperscript{\normalfont#1}}}
\renewcommand\AB@affilnote[1]{\expandafter\ifblank\expandafter{#1}{}{\textsuperscript{\normalfont#1}}}
\renewcommand{\maketitle}{%
  \begingroup
  \setlength{\parindent}{0pt}%
  \raggedright
  \vspace*{-12pt}%
  {\titlefont\exhyphenpenalty=0 \@title\par}%
  \vspace{12pt}%
  {\@author\par}%
  \ifx\@authornote\@empty\else
    \vspace{5pt}%
    {\fontsize{9}{11.5}\selectfont\@authornote\par}%
  \fi
  \vspace{12pt}%
  \endgroup}
\makeatother

\renewenvironment{abstract}
  {\par\begingroup\small\noindent\textbf{Abstract}\par\nobreak\vspace{3pt}\noindent\ignorespaces}
  {\par\endgroup\vspace{5pt}}

\hypersetup{
  pdftitle={GeoGuesser Bench: Benchmarking Vision-Language Agents on Interactive Geolocation},
  pdfauthor={Huanzhi Mao, Yichuan Wang, Michael Koo, Wenyuan Huang, Huy Doan, Jinghan Xia, Ziyao Wang, Joseph E. Gonzalez},
  pdfsubject={Interactive visual geolocation benchmark for vision-language agents},
  pdfkeywords={GeoGuesser Bench, visual geolocation, vision-language models, agents, exploration, Street View},
  bookmarksnumbered=true}

\title{\geobench{}: Benchmarking Vision--Language Agents on Interactive Geolocation}

\author[$*$]{Huanzhi~Mao}
\author[$*$]{Yichuan~Wang}
\author[ ]{Michael~Koo}
\author[ ]{Wenyuan~Huang}
\author[ ]{Huy~Doan}
\author[ ]{Jinghan~Xia}
\author[ ]{Ziyao~Wang}
\author[ ]{Joseph~E.~Gonzalez}
\affil[ ]{\mbox{University of California, Berkeley}}
\authornote{$*$\,Equal contribution.\quad Correspondence to: \texttt{huanzhimao@berkeley.edu}.}

\begin{document}

\maketitle
\setcounter{footnote}{1} 

\thispagestyle{firstpagestyle}

\begin{abstract}
Vision--language model (VLM) agents that act in the open world need two abilities: world knowledge to interpret what they see, and exploration to find what they cannot yet see.
We study how agents combine the two through interactive visual geolocation.
\geobench{} contains 282 tasks in an offline Street View environment where agents rotate, zoom, and move between real panoramas.
Its three task types shift the balance from knowledge to exploration: agents (1) report the coordinates of their starting location, (2) name the enclosing street, city, state, or country, or (3) find, read, and count nearby objects.
The coordinate tasks in (1) also include \numCompetition{} real GeoGuessr World Championship rounds, whose expert guesses provide a human reference.
We evaluate 23 VLMs: the best accuracy is 62.5\% on place naming and 35.0\% on object finding, and the best agent averages 4352.4 GeoGuessr points on the championship rounds, against \humanScore{} for the better expert in each round.
Beyond these headline numbers, world knowledge and exploration are only weakly correlated across models: agents that localize well do not necessarily explore well. To understand where agents fail, we categorize their errors into vision, navigation, and tool-use failures.
Further ablations show that the strongest agents, such as Claude Opus 4.8, still perform well with limited views, while weaker agents degrade; customizing the environment thus adjusts task difficulty.
The environment's verifiable rewards and restorable state also support training, and our Planner--Worker--Compressor harness cuts planner tokens by 57--61\% by compacting the interaction history.
We hope \geobench{} and its environment serve as a high-quality foundation for simulating real-world exploration at scale.
\end{abstract}

\section{Introduction}
\label{sec:introduction}

AI agents are moving from the screen into the physical world.
Success in the physical world requires more than recognizing what is already visible.
Agents need two abilities that are easy to conflate: \emph{knowing}, the world knowledge needed to interpret an observation, and \emph{looking}, the exploration needed to gather missing evidence by deciding where to look, how to move, and when enough has been seen.
A model may recognize an architectural style, a road marking, or a script from a single image, yet fail when the decisive clue lies around a corner.
Building such agents requires interactive testbeds that are grounded in the real world yet run digitally, at scale, and reproducibly, since physical deployment is too slow, costly, and uncontrolled to serve as a benchmark.

Current evaluations rarely exercise both abilities together.
Vision--language models (VLMs) increasingly act through tools~\citep{yang2023mmreact,liu2023llavaplus,suris2023vipergpt}, but static visual benchmarks pose questions about a fixed image, and agent benchmarks mostly target function calling and digital workflows~\citep{patil2025bfcl,trivedi2024appworld} or configurable puzzle, navigation, and manipulation environments~\citep{wang2026visgym}.
It thus remains unclear whether models with strong visual world knowledge can also explore a real environment for evidence that is not initially visible.

Visual geolocation from Street View is a natural testbed for this question.
Street View meets these requirements: its imagery carries the diversity and ambiguity of the real world, yet an agent can rotate, zoom, and move between panoramas without deploying hardware.
Success can hinge on knowledge, such as recognizing architecture, signage, or vegetation, or on exploration, such as walking to a street sign or storefront.
The game GeoGuessr has turned the task into a competitive sport~\citep{haas2024pigeon,plonkit} whose top players offer a strong human reference.
Most geolocation benchmarks, however, score a single prediction from a fixed image~\citep{haas2024pigeon}, and interactive geolocation work~\citep{zheng2026wander} remains centered on localization: it does not ask agents to seek out other information in the scene or compare them with expert players.

We introduce \geobench{}, 282 tasks in an interactive Street View environment in which agents explore through tools and return answers that are verified automatically.
Three task types shift the balance from knowing to looking.
\typeOne{} asks for the coordinates of the starting location and rewards world knowledge.
\typeTwo{} asks for the enclosing street, city, state, or country, whose clues may require turning or moving to bring into view.
\typeThree{} asks agents to find, read, or count objects that are not visible from the start and must be actively discovered.
The coordinate tasks include \numCompetition{} World Championship rounds, with the finalists' guesses as a human reference.

We evaluate 23 VLMs, and four findings stand out.
First, knowledge-heavy and exploration-heavy tasks rank models differently: GPT-5.4 ranks third on one coordinate split but ties for 18th on object finding.
Second, even the best agent still falls short of human experts on the championship rounds, averaging 4352.4 points against \humanScore{} for the experts.
Third, stronger agents hold up when the environment gets harder, while weaker agents break down: when each view is cut to its top or bottom half, Claude Opus 4.8 retains 98\% of its score, whereas Gemini 3.1 Pro retains only 33\%.
Finally, a failure case study of four models shows that navigational reasoning, rather than visual recognition or tool use, is the most common cause of failure, appearing in 60--80\% of failed episodes on every task.

The environment is also designed for training.
It runs fully offline as a self-contained package, which makes it easy to set up and use, keeps observations from drifting as Street View imagery is updated, and closes shortcuts to reward hacking, since agents cannot look answers up.
Its rewards are verifiable, and its full state (panorama and camera pose) can be restored at any step, which makes branching rollouts from intermediate states~\citep{yi2026pivotrl} inexpensive.
Its difficulty is also configurable: restricting the agent's views or tools (e.g., removing movement) changes what evidence it can gather, and restricted views cost weaker agents far more than the strongest, which can support a training curriculum.

Long episodes are also costly: every step appends an image and text that the model rereads at every later step, so the input grows quickly with episode length.
We therefore introduce a simple context-management harness, Planner--Worker--Compressor, that records the exploration state in a short summary before each planning call while keeping selected key frames.
It prunes planner tokens by 57--61\% while largely preserving accuracy.

\textbf{Contributions.}
\begin{itemize}[leftmargin=2em]
    \item \textbf{Benchmark.} \geobench{}: 282 real-world Street View tasks that range from knowledge-heavy localization to exploration-heavy information seeking, with an expert reference from GeoGuessr World Championship rounds (Sections~\ref{sec:data_curation} and~\ref{sec:eval_setup}).
    \item \textbf{Environment.} An offline, configurable environment with verifiable rewards and restorable state for evaluating and training agents, and a Planner--Worker--Compressor harness that reduces planner tokens by 57--61\% (Section~\ref{sec:method}).
    \item \textbf{Findings.} An evaluation of 23 VLMs, with ablations and a failure analysis, showing that knowledge and exploration rank models differently, that agents still trail human experts, that stronger agents are more robust to restricted views, and that navigational reasoning is the main failure mode (Section~\ref{sec:results}).
\end{itemize}

\section{Related Work}
\label{sec:related_work}

\paragraph{Visual geolocation}
Most geolocation methods predict a location from a single image, whether by classifying over geographic cells~\citep{weyand2016planet,Theiner2021Interpretablesamanticphotogeolocalization}, fusing image and segmentation features~\citep{pramanick2022worldimagetransformerbasedgeolocalization}, aligning images with GPS coordinates contrastively~\citep{cepeda2023geoclipclipinspiredalignmentlocations}, combining geographic priors with supervised training~\citep{haas2024pigeon}, augmenting multimodal LLMs with retrieval~\citep{zhou2024img2loc,jia2024g3effectiveadaptiveframework}, fusing semantics with hierarchical geographic embeddings~\citep{daruna2026geosurgegeolocalizationusingsemantic}, or re-ranking candidates by distance~\citep{jia2025georankerdistanceawarerankingworldwide}.
Agent-based methods add retrieval and OCR tools but still reason over one fixed view~\citep{li2026locationagenthierarchicalagentimage}.
WanderBench~\citep{zheng2026wander} makes geolocation interactive by organizing panoramas into navigable graphs.
\geobench{} also asks agents to act, but additionally tests information seeking beyond localization (\typeThree{}) and provides an expert-player reference.

\paragraph{Interactive and embodied benchmarks}
StreetLearn~\citep{mirowski2018learning} and Touchdown~\citep{chen2019touchdown} evaluate goal-driven and instruction-following navigation in Street View, where success depends on reaching a place rather than on recognizing where one is.
VisGym~\citep{wang2026visgym} evaluates VLMs across configurable puzzle, navigation, and manipulation environments, and Agent-X~\citep{ashraf2025agentx} evaluates multi-step visual reasoning with tools.
Tool-use benchmarks~\citep{patil2025bfcl,trivedi2024appworld,ning2024wtueval} test function calling in text or digital environments, and vision--language--action models target physical control~\citep{ma2024vla}.
\geobench{} complements these with global real-world imagery and answers that require both world knowledge and exploration.

\paragraph{Agent frameworks and context management}
MM-REACT~\citep{yang2023mmreact}, LLaVA-Plus~\citep{liu2023llavaplus}, and ViperGPT~\citep{suris2023vipergpt} pair language or vision--language controllers with vision tools, and ReAct~\citep{yao2023react} interleaves reasoning with actions in a closed loop.
Toolformer~\citep{schick2023toolformer} and Gorilla~\citep{patil2023gorilla} train models to call APIs; see \citet{qin2023tool} for a survey of tool learning.
Long interaction histories remain a bottleneck: VisGym finds that VLMs perform worse with unbounded histories than with truncated ones~\citep{wang2026visgym}.
Our Planner--Worker--Compressor harness (Section~\ref{sec:method}) addresses this by summarizing the textual history while keeping selected visual observations.

\section{Benchmark Construction}
\label{sec:data_curation}

\begin{figure*}[t]
    \centering
    \includegraphics[width=\textwidth]{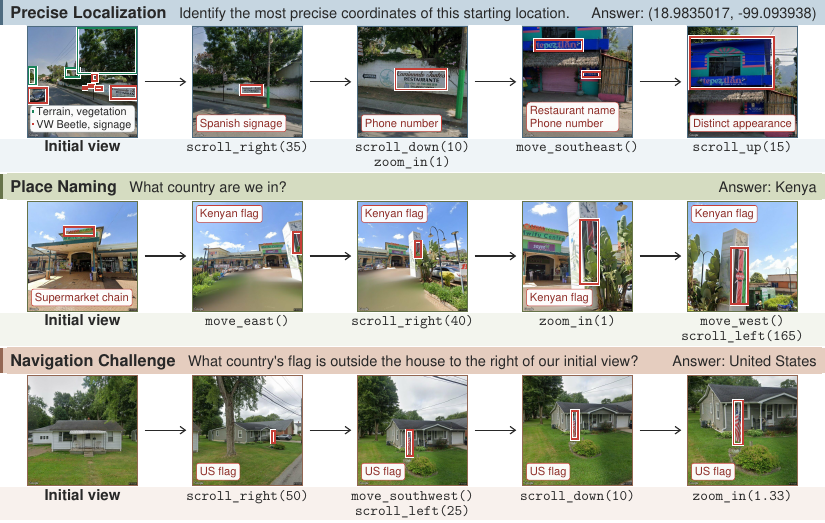}
    \caption{\textbf{Example \geobench{} episodes.} Read left to right; calls below each new view specify the transition. Boxes mark natural (green) and human-made (red) clues. From top: \typeOne{}, \typeTwo{}, and \typeThree{}. Prompts are abbreviated or illustrative.}
    \label{fig:type_examples}
    \label{fig:tool_demonstration}
\end{figure*}

\geobench{} contains 282 entries, each pairing a Street View starting location with a question and a reference answer (Figure~\ref{fig:type_examples}). All entries are solved in the same interactive environment (Section~\ref{sec:eval_setup}), but they differ in how much they draw on world knowledge versus exploration: by design, \typeOne{} depends mainly on knowledge, \typeThree{} mainly on exploration, and \typeTwo{} on both. Section~\ref{sec:results} tests whether agents actually solve them this way.

\subsection{Task Types}
\label{subsec:task_types}

\paragraph{\typeOne{} (\numTypeOne{} + \numCompetition{} entries)}
The agent reports the latitude and longitude of its starting location, as in GeoGuessr. Precise answers usually require combining many individually ambiguous cues, both natural (vegetation, soil, terrain) and human-made (signage, road markings, architecture). The task has two splits: the \emph{Synthetic} split, \numTypeOne{} real locations that we selected ourselves, and the \emph{\typeCompetition{}} split, \numCompetition{} rounds of professional GeoGuessr play with recorded expert guesses (Section~\ref{subsec:curation}).

\paragraph{\typeTwo{} (\numTypeTwo{} entries)}
The agent names the region containing the starting location at the granularity the question specifies: street (11 entries), city or town (31), state or province (11), or country (27). Each location has regional indicators nearby, such as street signs, the language of written text, or architectural style, but the agent may need to turn or move to bring them into view. The task thus combines cultural knowledge with moderate navigation.

\paragraph{\typeThree{} (\numTypeThree{} entries)}
The agent answers a question about a specific object or feature near the start. Question types include \findingTypeThree{}, which asks for a property of a target, such as a business name, the text on a sign, or a color, and \countingTypeThree{}, which asks how many instances of an object lie in an area or along a route. Many prompts give relative directions (e.g., ``head towards the stoplight''). No target is visible in the initial view, so every question requires changing the viewpoint, and many require moving to other panoramas. The task therefore demands the most precise navigation and, on average, the least world knowledge.

\subsection{Curation}
\label{subsec:curation}

\begin{figure*}[t]
    \centering
    \includegraphics[width=0.85\textwidth]{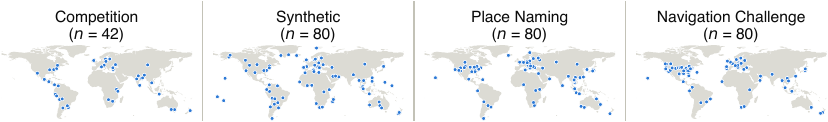}
    \caption{\textbf{Starting locations of each split.} \typeCompetition{} and Synthetic are the two \typeOne{} splits; \typeTwo{} and \typeThree{} favor urban and suburban areas (Section~\ref{subsec:curation}). Empty regions largely reflect scarce or absent Street View imagery, as in mainland China and much of Russia and Africa.}
    \label{fig:global_distribution}
\end{figure*}

\paragraph{Locations}
We selected the Synthetic locations manually to span diverse countries, climates, and settings, from city centers to rural roads. \typeTwo{} and \typeThree{} locations favor urban and suburban areas, where cultural signals are strong and objects are distinct (Figure~\ref{fig:global_distribution}). For the Synthetic split and \typeTwo{}, whose answers are properties of the location itself, we excluded locations whose initial view shows a famous landmark that a model could recognize without exploring. Together with the offline environment (Section~\ref{sec:eval_setup}), this filtering is designed so that answers must come from the scene rather than from recalling a landmark or searching externally.

\paragraph{Questions and accepted answers}
For each \typeTwo{} location, we chose a target region and identified visual clues near the start that point to it; where necessary, we changed the granularity (e.g., from city to street) or moved the start within the region so that the clues fit the target. For each \typeThree{} location, we chose a target that exploration can reliably reach and wrote a question whose answer is a verifiable fact in the scene; we confirmed that the target is not visible in the initial view and that the question can be answered within the 40-step budget. Each \typeTwo{} and \typeThree{} question specifies an answer format (e.g., ``Give only the number as an integer'') and lists accepted answers that cover equivalent forms, such as ``Ave'' and ``Avenue'' or ``Czechia'' and ``Czech Republic.'' For \typeOne{}, the answer is the starting coordinate.

\paragraph{Competition split}
The \numCompetition{} \typeCompetition{} locations are rounds from the Grand Finals of the GeoGuessr World Cup in 2023 and 2024 (renamed the GeoGuessr World Championship in 2025). For each round, we record the location together with the distance and score of the better of the two players' guesses, which form the human reference (Section~\ref{sec:eval_setup}). Because these rounds were broadcast publicly, models may have seen them during pretraining; the Synthetic split, which does not come from public competition, serves as a check against such memorization (Section~\ref{subsec:agent_performance}).

\section{Evaluation Setup}
\label{sec:eval_setup}

\paragraph{Environment}
The environment exposes Google Street View panoramas to the agent through tools (Figure~\ref{fig:tool_demonstration}). An episode starts at the entry's starting location with the camera facing north, level, at the default zoom (a $90^\circ$ horizontal field of view). The prompt contains no image: the agent sees the scene only through tools that return a $1920\times1080$ rendering of the current view. Renderings omit Street View's interface overlays, including the address box, road labels, and navigation arrows, so the interface never reveals the location. The agent has the following tools: \texttt{scroll\_*} tools rotate (left, right) or tilt (up, down) the camera, and \texttt{zoom\_in} and \texttt{zoom\_out} change magnification, each by an amount the agent chooses; and \texttt{move\_*} tools (\texttt{move\_north}, \texttt{move\_northeast}, \ldots) step to the adjacent panorama in a compass direction, if one exists. Camera tools return the current view.

\paragraph{Offline package}
The environment is a self-contained package that runs fully offline. The panoramas and panorama links reachable from each entry are bundled with the benchmark, so episodes make no network requests and observations do not drift as Street View imagery is updated. Running offline also closes shortcuts to reward hacking: agents cannot query map services, reverse-geocode their position, or search the web, and tools return only rendered views or movement status, never coordinates, addresses, or reference answers. The task set, tool set, step budget, and observation settings are configurable, and only the web-search ablation (Section~\ref{subsec:ablation_studies}) enables internet access. We will release the package with the benchmark.

\paragraph{Episodes and models}
A step is one model response, which may contain several tool calls. An episode ends when the agent replies without calling a tool, and that reply is its answer; the agent may take at most 40 tool-calling steps. The system prompt describes how answers are scored, asks for a final reply with an \texttt{answer} field and a brief \texttt{context} justification, and permits ``I do not know'' as an answer. An agent that uses all 40 steps receives one final turn in which it is asked for its answer; we say that an episode \emph{ends at the step limit} only if the agent still gives no answer in this turn. We evaluate the 23 VLMs in Table~\ref{tab:results-summary} through their function-calling interfaces with identical prompts, tools, and budget. Every entry counts toward its task's result: an episode that ends at the step limit, ends in an error, or yields no parseable answer scores 0 on \typeOne{} and counts as incorrect on \typeTwo{} and \typeThree{}.

\paragraph{Scoring}
For \typeOne{}, we parse the first valid latitude--longitude pair in the final reply and compute its geodesic distance $d$ (km) to the ground truth on the WGS-84 ellipsoid. GeoGuessr does not publish its scoring rule, but prior work and player analyses report an exponential decay~\citep{haas2024pigeon,Latb}, $S(d) = 5000\exp(-10d/D)$, where $D$ is the haversine length of the diagonal of the smallest latitude--longitude rectangle containing the map ($D = 14{,}916.862$~km for the world map; Appendix~\ref{app:type1-scoring}). The score ranges from 0 to 5000. For \typeTwo{} and \typeThree{}, we report accuracy over all 80 entries of each task; a reply is correct only if its \texttt{answer} field strictly matches one of the accepted answers, which include equivalent aliases (Appendix~\ref{app:answer-matching}).

\paragraph{Human reference}
On the \typeCompetition{} split, the better player's guess in each round averages \humanScore{} points, a score-equivalent distance (Appendix~\ref{app:type1-scoring}) of 70.2~km. This is a reference point, not a matched comparison: players had one minute per round and placed their guesses on an interactive map, whereas agents have up to 40 steps but must state coordinates without a map. Taking the better of two guesses per round also makes the reference optimistic relative to a single expert.

\section{Results}
\label{sec:results}

\subsection{Agent Performance}
\label{subsec:agent_performance}

\begin{table*}[t]
  \centering
  \small
  \setlength{\tabcolsep}{2.5pt}
  \caption{\textbf{Results on \geobench{}.} For the \typeCompetition{} and Synthetic splits of \typeOne{} (Section~\ref{subsec:task_types}), cells show the mean GeoGuessr score over all entries, with invalid responses scored 0, and the score-equivalent distance in km (Eq.~\ref{eq:score-equivalent-distance}). The human row averages the better of the two players' guesses in each \typeCompetition{} round (Section~\ref{sec:eval_setup}). Models are sorted by average rank, the mean of a model's ranks in the four result columns (1 is best; ties share the mean rank). \textbf{Bold} marks the best model in each column.}
  \label{tab:results-summary}
  \begin{tabular*}{\textwidth}{@{\extracolsep{\fill}}lcccc@{}}
    \toprule
    & \multicolumn{2}{c}{\textbf{\typeOne} (Score\,$|$\,Dist)} & \taskNameCell{\textbf{\typeTwo}} & \taskNameCell{\textbf{\typeThree}} \\
    \cmidrule(lr){2-3}
    \textbf{Model} &
    \textbf{\makecell[b]{\typeCompetition}} &
    \textbf{\makecell[b]{Synthetic}} &
    (\%) &
    (\%) \\
    \midrule
  Human experts        & \humanScore{}\,$|$\,70.2 & -- & -- & -- \\
    \midrule
  Claude Opus 4.8      & \textbf{4352.4\se{128}\,$|$\,206.9} & \textbf{4549.5\se{83}\,$|$\,140.8} & 60.00\se{3.6} & 26.25\se{3.3} \\
  Claude Opus 4.7      & 4306.4\se{146}\,$|$\,222.8 & 4501.4\se{96}\,$|$\,156.7 & 56.25\se{3.6} & 31.25\se{3.4} \\
  GPT-5.5              & 3815.1\se{182}\,$|$\,403.5 & 3467.8\se{161}\,$|$\,545.8 & \textbf{62.50\se{3.5}} & 32.50\se{3.4} \\
  Claude Fable 5       & 2821.1\se{248}\,$|$\,853.7 & 3164.2\se{175}\,$|$\,682.5 & \textbf{62.50\se{3.5}} & \textbf{35.00\se{3.5}} \\
  Qwen3.7-Plus         & 3964.4\se{133}\,$|$\,346.2 & 3996.0\se{104}\,$|$\,334.4 & 42.50\se{3.6} & 23.75\se{3.1} \\
  GPT-5.4 Pro          & 2503.8\se{244}\,$|$\,1031.7 & 2667.4\se{176}\,$|$\,937.3 & 60.00\se{3.6} & \textbf{35.00\se{3.5}} \\
  GPT-5.4              & 3923.1\se{122}\,$|$\,361.8 & 4221.1\se{85}\,$|$\,252.6 & 50.00\se{3.6} & 12.50\se{2.4} \\
  GLM-5V-Turbo         & 3138.2\se{200}\,$|$\,694.8 & 3859.8\se{131}\,$|$\,386.1 & 46.25\se{3.6} & 15.00\se{2.6} \\
  Claude Opus 4.6      & 3023.9\se{231}\,$|$\,750.2 & 3363.5\se{165}\,$|$\,591.4 & 45.00\se{3.6} & 18.75\se{2.9} \\
  MiniMax-M3           & 3061.5\se{205}\,$|$\,731.7 & 2905.8\se{159}\,$|$\,809.6 & 33.75\se{3.4} & 17.50\se{2.8} \\
  Qwen3.6-Plus         & 3045.5\se{173}\,$|$\,739.5 & 2956.4\se{141}\,$|$\,783.8 & 36.25\se{3.5} & 15.00\se{2.6} \\
  GPT-5.5 Pro          & 1163.4\se{211}\,$|$\,2175.0 & 1556.1\se{168}\,$|$\,1741.2 & 47.50\se{3.6} & 31.25\se{3.4} \\
  Gemini 3 Flash       & 2851.0\se{240}\,$|$\,838.0 & 2493.3\se{183}\,$|$\,1038.0 & 38.75\se{3.6} & 18.75\se{2.9} \\
  Claude Sonnet 4.6    & 2649.1\se{245}\,$|$\,947.5 & 2883.8\se{174}\,$|$\,820.9 & 47.50\se{3.6} & 12.50\se{2.4} \\
  Gemini 3.5 Flash     & 1698.3\se{233}\,$|$\,1610.7 & 2184.0\se{180}\,$|$\,1235.5 & 46.25\se{3.6} & 17.50\se{2.8} \\
  Gemini 3.1 Pro       & 593.6\se{164}\,$|$\,3178.8 & 1124.8\se{153}\,$|$\,2225.3 & 47.50\se{3.6} & 30.00\se{3.4} \\
  Kimi K2.5            & 2016.2\se{228}\,$|$\,1354.8 & 2717.1\se{174}\,$|$\,909.7 & 42.50\se{3.6} & 15.00\se{2.6} \\
  Kimi K2.6            & 1667.9\se{229}\,$|$\,1637.7 & 2100.3\se{177}\,$|$\,1293.8 & 50.00\se{3.6} & 13.75\se{2.5} \\
  Qwen3.5-397B-A17B    & 2221.5\se{209}\,$|$\,1210.1 & 2467.0\se{160}\,$|$\,1053.8 & 28.75\se{3.3} & 8.75\se{2.1} \\
  GPT-5.2              & 523.9\se{146}\,$|$\,3365.1 & 1243.2\se{151}\,$|$\,2076.1 & 33.75\se{3.4} & 17.50\se{2.8} \\
  Grok 4.3             & 805.8\se{173}\,$|$\,2722.9 & 1546.8\se{162}\,$|$\,1750.1 & 32.50\se{3.4} & 5.00\se{1.6} \\
  Qwen3.5-122B-A10B    & 145.7\se{78}\,$|$\,5274.1 & 807.2\se{126}\,$|$\,2720.3 & 12.50\se{2.4} & 6.25\se{1.8} \\
  Qwen3.5-35B-A3B      & 252.4\se{83}\,$|$\,4454.4 & 364.8\se{89}\,$|$\,3905.0 & 10.00\se{2.2} & 3.75\se{1.4} \\
  \bottomrule
  \end{tabular*}
\end{table*}

\paragraph{\typeOne{}}
No model reaches the human baseline on the \typeCompetition{} split (Table~\ref{tab:results-summary}). The best agent, Claude Opus 4.8, averages 4352.4 points (207~km) against the expert reference of \humanScore{} (70~km), and the runner-up trails it by only 46 points, well within their standard errors of about 200. The two \typeOne{} splits rank the 23 models almost identically (Spearman $\rho = 0.97$), and 19 models score higher on Synthetic than on \typeCompetition{}, so we see no sign that exposure to broadcast competition rounds inflates \typeCompetition{} scores.

Much of the spread between models reflects whether an agent answers at all rather than how accurately it localizes. Pooled over models, 45.7\% of \typeCompetition{} and 38.6\% of Synthetic episodes score zero (Table~\ref{tab:zero-causes}), mostly because the agent either ends at the step limit, using all 40 steps without answering even when prompted in the final turn (20.7\% and 19.1\% of episodes; Section~\ref{sec:eval_setup}), or abstains before the limit (19.0\% and 14.6\%). Both are failures to commit to an answer rather than inaccurate answers: the system prompt permits ``I do not know,'' but it also describes the scoring rule, under which any guess scores above zero, so declining to guess is never optimal. When agents do answer, they are often accurate (Table~\ref{tab:valid-answers}): on the Synthetic episodes they answer, GPT-5.2 and Grok~4.3, whose zeros are all abstentions, average 4324 and 4419 points, more than GPT-5.4 (4221), which answers all 80. This conditional score is only a diagnostic, since each model answers a different subset of episodes.

\begin{figure*}[t]
    \centering
    \includegraphics[width=0.85\textwidth]{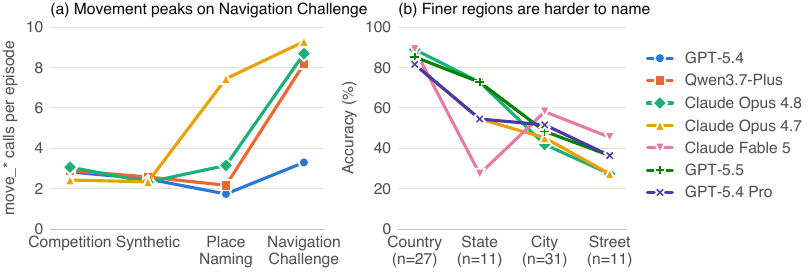}
    \caption{\textbf{Exploration and region granularity.} (a) Mean \texttt{move\_*} calls per completed episode for the four strongest models by mean \typeOne{} score: all four move little on \typeOne{}, and three move far more on \typeThree{}. (b) Accuracy of the five strongest \typeTwo{} models by the granularity the question asks for. Claude Fable 5's dip at the state level comes from the step budget: all eight state questions it missed ended at the step limit. Tables~\ref{tab:movement-calls} and~\ref{tab:type2-granularity} list the values.}
    \label{fig:exploration-granularity}
\end{figure*}

The strongest \typeOne{} agents also explore little: the four best by mean \typeOne{} score average at most 3.1 \texttt{move\_*} calls per episode on this task, yet three of them move 8.2--9.3 times per \typeThree{} episode (Figure~\ref{fig:exploration-granularity}a). Section~\ref{subsec:ablation_studies} tests whether movement is needed at all.

\paragraph{\typeTwo{}}
The best models name the correct region in 62.5\% of questions. Accuracy falls with granularity (Figure~\ref{fig:exploration-granularity}b): the five strongest models identify the country in 81.5--88.9\% of cases but the street in only 27.3--45.5\%. Running out of steps also costs accuracy: 16.1\% of all \typeTwo{} episodes end at the step limit.

\paragraph{\typeThree{}}
This is the hardest task: no model exceeds 35.0\%. Agents also work longest here with 19 of the 23 models taking more steps on \typeThree{} than on either the Synthetic split or \typeTwo{} (Table~\ref{tab:step-counts}), and 28.5\% of \typeThree{} episodes end at the step limit, accounting for 35.3\% of errors.

\paragraph{Rankings across tasks}
Rankings on one task only partly predict the others. Across the 23 models, \typeOne{} scores correlate weakly with \typeThree{} accuracy (Spearman $\rho = 0.39$ for both splits) and moderately with \typeTwo{} accuracy ($\rho = 0.48$ and $0.55$), whereas \typeTwo{} and \typeThree{} correlate more strongly ($\rho = 0.70$). With 23 models these estimates are imprecise (a Fisher-$z$ 95\% interval for $\rho = 0.39$ spans $-0.02$ to $0.69$), but individual rank reversals are large. GPT-5.4 is third on the Synthetic split but ties for 18th on \typeThree{}, where it takes only 10.7 steps per episode, consistent with too little exploration; conversely, Claude Fable 5 ties for first on \typeTwo{} and \typeThree{} but ranks 11th on \typeCompetition{}, partly because it exhausts the budget in 15 of 42 episodes.

\subsection{Failure Mode Analysis}
\label{subsec:failure_modes}

\begin{figure}[t]
    \centering
    \begin{minipage}[t]{0.40\linewidth}
        \vspace{0pt}
        \centering
        \small
        \captionsetup{justification=raggedright}
        \captionof{table}{\textbf{Failure modes by task.} Share of failed episodes with each mode, pooled over the four annotated models; an episode can have several modes. \textbf{Bold}: most common.}
        \label{tab:fm_type_distribution}
        \setlength{\tabcolsep}{3pt}
\begin{tabular}{@{}lccc@{}}
    \toprule
    \textbf{Task} & \makecell{\textbf{FM1}\\vision} & \makecell{\textbf{FM2}\\navigation} & \makecell{\textbf{FM3}\\tool use} \\
    \midrule
    Comp.   & 0.136 & \textbf{0.780} & 0.152 \\
    Synth.  & 0.154 & \textbf{0.802} & 0.123 \\
    \parbox[c]{52pt}{\raggedright\strut\typeTwo{}\strut} & 0.326 & \textbf{0.665} & 0.035 \\
    \parbox[c]{52pt}{\raggedright\strut\typeThree{}\strut} & 0.351 & \textbf{0.601} & 0.145 \\
    \bottomrule
\end{tabular}

    \end{minipage}\hfill
    \begin{minipage}[t]{0.57\linewidth}
        \vspace{0pt}
        \centering
        \includegraphics[width=0.85\linewidth]{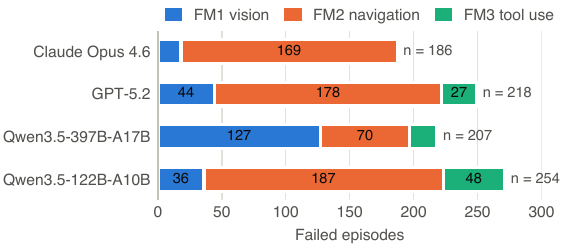}
        \caption{\textbf{Failure modes by model.} Failed episodes assigned each mode, pooled over the four task splits; $n$ is the number of failed episodes, and an episode can have several modes.}
        \label{fig:fm_counts_horizontal}
    \end{minipage}
\end{figure}
To characterize errors, we assigned each failed episode of four models (Claude Opus 4.6, GPT-5.2, and Qwen3.5 at 122B and 397B parameters) one or more of three failure modes (Table~\ref{tab:failure-modes}). A \typeOne{} episode counts as failed if it scores worse than 4000 points (333~km), or gave no valid answer; a \typeTwo{} or \typeThree{} episode fails if its answer does not match the ground truth for that question. We labeled 160 of the 865 failed episodes manually (10 per model and task split) and used them as examples for LLM-based classifier agents, which labeled the remaining 705. \textbf{FM1, Vision Failure}, covers errors in interpreting visual input or applying pretrained knowledge, such as hallucinated landmark recognition, misclassified cues, or fixation on distractors (Figures~\ref{fig:fm_1_figure} and~\ref{fig:fm_2_figure}). \textbf{FM2, Navigational Reasoning Failure}, covers errors in spatial reasoning and planning, such as inefficient routes, failing to follow the prompt's directions or to replan, and losing track of objects or explored areas (Figures~\ref{fig:fm_3_figure} and~\ref{fig:fm_4_figure}). \textbf{FM3, Tool-use Failure}, covers malformed, invalid, or unsupported tool requests, such as misreading a documented tool or calling one that does not exist, which cause API errors, refusals to engage with the task, or corrupted context (Figure~\ref{fig:fm_5B_figure}). Figure~\ref{fig:fm_examples} in Appendix~\ref{app:additional} shows three of these examples side by side.

Navigational reasoning failure is the most common mode on every task split, appearing in 60--80\% of failed episodes (Table~\ref{tab:fm_type_distribution}), and it accounts for most failures of three of the four models (Figure~\ref{fig:fm_counts_horizontal}). Within the Qwen3.5 family, the larger model has fewer tool-use failures (21 versus 48) but more vision failures (127 versus 36), possibly due to longer interactions with the environment (taking more steps on every task; Table~\ref{tab:step-counts}), which bloats context with redundant visual evidence.

\subsection{How Agents Gather Evidence}
\label{subsec:ablation_studies}

\begin{table*}[t]
\centering
\small
\setlength{\tabcolsep}{4pt}
\caption{\textbf{Ablation results across all conditions.} Mean GeoGuessr score (0--5000) per model on the two \typeOne{} splits, Competition (C, $n=\numCompetition$) and Synthetic (S, $n=\numTypeOne$), with invalid responses assigned zero. \emph{80 Steps}: step limit raised from 40 to 80, navigation enabled. \emph{No Movement}: \texttt{move\_*} tools removed, look-and-zoom only, 40 steps. \emph{Top/Bottom Half}: only the named half of each $1920\times1080$ capture retained, 40 steps, navigation enabled. \emph{Web Search}: \texttt{search\_engine\_query} and \texttt{fetch\_url\_content} added, full images, 40 steps. Models are sorted by average rank over the ten columns; best value per column in bold. GPT-5.5's Web Search cell on Synthetic averages 79 entries because one question was not attempted (1809 if it is counted as zero). Claude Opus 4.6's Top Half, Bottom Half, and Web Search runs used OpenRouter; its other runs used the provider's API.}
\label{tab:ablations-combined}
\begin{tabular*}{\textwidth}{@{\extracolsep{\fill}}l*{10}{c}@{}}
\toprule
\multirow{2}{*}{\textbf{Model}} & \multicolumn{2}{c}{\textbf{80 Steps}} & \multicolumn{2}{c}{\textbf{No Movement}} & \multicolumn{2}{c}{\textbf{Top Half}} & \multicolumn{2}{c}{\textbf{Bottom Half}} & \multicolumn{2}{c}{\textbf{Web Search}} \\
\cmidrule(lr){2-3} \cmidrule(lr){4-5} \cmidrule(lr){6-7} \cmidrule(lr){8-9} \cmidrule(lr){10-11}
 & C & S & C & S & C & S & C & S & C & S \\
\midrule
Claude Opus 4.8   & \textbf{4594} & \textbf{4805} & \textbf{4646} & \textbf{4556} & \textbf{4368} & \textbf{4386} & \textbf{4369} & \textbf{4313} & \textbf{4551} & \textbf{4380} \\
GLM-5V-Turbo      & 3520 & 4089 & 4249 & 4356 & 3061 & 3960 & 3093 & 3480 & 3544 & 4022 \\
Qwen3.7-Plus      & 3884 & 3930 & 3686 & 3998 & 2925 & 3171 & 3271 & 3097 & 3080 & 3339 \\
Claude Opus 4.6   & 3708 & 3741 & 3749 & 3018 & 2527 & 2583 & 1928 & 2879 & 2688 & 3069 \\
Claude Sonnet 4.6 & 2791 & 3193 & 3626 & 4203 & 2337 & 2922 & 2446 & 2707 & 3206 & 2767 \\
GPT-5.5           & 4407 & 4391 & 4100 & 4178 & 2212 & 2305 & 1403 & 1553 & 2388 & 1832 \\
Kimi K2.5         & 3038 & 3167 & 2668 & 3119 & 2258 & 2413 & 718 & 1819 & 3999 & 3979 \\
Gemini 3.5 Flash  & 2744 & 3228 & 3939 & 4019 & 948 & 2123 & 938 & 1179 & 0 & 125 \\
Qwen3.5-397B      & 2175 & 2149 & 1661 & 1861 & 1871 & 2235 & 1556 & 1147 & 2444 & 2662 \\
Kimi K2.6         & 1218 & 1804 & 781 & 2161 & 471 & 1599 & 856 & 1185 & 238 & 507 \\
Gemini 3.1 Pro    & 1729 & 2473 & 1939 & 2532 & 0 & 1062 & 0 & 437 & 119 & 312 \\
GPT-5.2           & 767 & 1295 & 572 & 1544 & 604 & 902 & 112 & 667 & 1696 & 2209 \\
Grok 4.3          & 650 & 1400 & 578 & 623 & 407 & 1133 & 213 & 820 & 835 & 1022 \\
\bottomrule
\end{tabular*}
\end{table*}

We probe how agents gather evidence for \typeOne{} with five single-factor interventions on 13 models (Table~\ref{tab:ablations-combined}): an 80-step limit, no \texttt{move\_*} tools, only the top or bottom half of each view, and web search. We compare each with the 40-step scores in Table~\ref{tab:results-summary}.

\paragraph{Step budget and movement}
A larger budget and the removal of movement both raise scores for most models, largely because agents run out of steps less often. With 80 steps, 9 of the 13 models improve on each split, and the share of episodes that end at the step limit falls from 19.1\% to 7.4\% (pooled over models and splits). Three of the four models that do not improve almost never end at the 40-step limit in the main setting.

Removing movement at the original 40-step budget raises scores over the main setting for 9 of 13 models on \typeCompetition{} and 10 on Synthetic, and over the 80-step condition for 6 and 7. Episodes become shorter (12.3 versus 20.0 steps on average) and rarely reach the limit (4.5\%, versus 19.1\% at 40 steps and 7.4\% at 80); Claude Opus 4.8 scores 4646 on \typeCompetition{} without moving at all, more than with movement at either budget. For the strongest agents, a single panorama thus suffices for strong \typeOne{} performance, and for most models the main cost of movement is the budget it consumes. The exceptions are two models that tend to abstain, which score lower without movement than with it at either budget and abstain more often.

\paragraph{Image crops}
Keeping only the top or bottom half of each view lowers scores for 10 to 13 of the 13 models, depending on the half and split. Stronger models are more robust: averaged over both crops and splits, the fraction of its main-setting score that a model retains (33--98\%) rises with its mean main-setting \typeOne{} score (Spearman $\rho = 0.73$; Fisher-$z$ 95\% interval 0.29 to 0.91), with GPT-5.5 the clearest exception (third in the main setting, it retains 52\%). Neither half is consistently more informative. Although the top half beats the bottom half for 12 of 13 models on Synthetic and for only 7 on \typeCompetition{}, every model tilts toward the missing half more often than away from it, which costs steps. The share of episodes ending at the step limit rises from 19.1\% to 28.1\% (top) and 30.2\% (bottom).

\paragraph{Web search}
Web search improves 7 of 13 models on \typeCompetition{} but only 4 on Synthetic. It helps agents that would otherwise abstain: Kimi K2.5 nearly doubles its \typeCompetition{} score (2016 to 3999), abstaining in 4 rather than 47 episodes across both splits. It hurts agents that spend their budget searching: Gemini 3.5 Flash scores zero on all 42 \typeCompetition{} episodes. External information therefore does not reliably help; agents must still find useful clues and budget their steps.

\begin{figure}[t]
    \centering
    \begin{minipage}[b]{0.42\textwidth}
        \centering
        \includegraphics[width=0.85\linewidth]{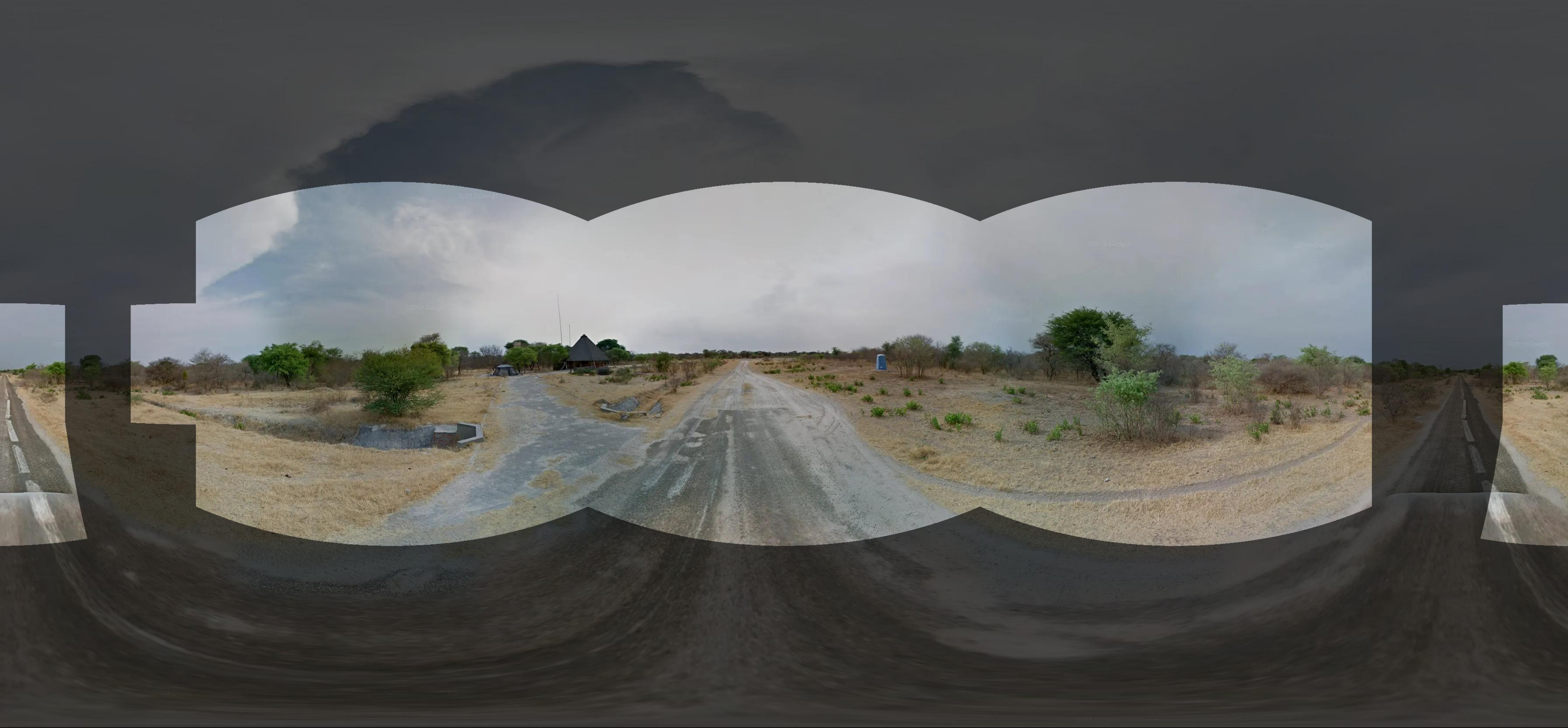}
    \end{minipage}\hfill
    \begin{minipage}[b]{0.55\textwidth}
        \caption{\textbf{Visual coverage within one panorama.} Bright regions fall within the reconstructed camera views of an example no-movement episode; dark regions were never viewed. The mask shows which directions were viewed, not what the agent attended to.}
        \label{fig:nomovement_figure}
    \end{minipage}
\end{figure}

\paragraph{Visual coverage}
Agents rarely inspect the full panorama. We measure coverage, the fraction of the viewing sphere covered by the union of an agent's camera views in a no-movement episode (Figure~\ref{fig:nomovement_figure}; Appendix~\ref{app:visual-context}). Across ten models, about two-thirds of episodes with any tool call never change pitch, and most end below the 0.49 of the sphere covered by one full horizontal sweep at the default field of view (Figure~\ref{fig:coverage-performance}). High scores do not require broad coverage: Claude Opus 4.8's no-movement run above has a median coverage of 0.31 and never tilts the camera. One explanation is that many geographic cues, such as signage, road markings, vegetation, and facades, lie near the horizon, echoing the equator bias of people viewing $360^\circ$ panoramas~\citep{sitzmann2018saliency}.

\section{Mitigating Context Growth in Long Episodes}
\label{sec:method}

Each step appends an image and text that the agent rereads at every later step, so its context grows linearly with episode length and its total input grows quadratically. For Gemini~3 Flash on \typeThree{}, the median input rises from about 1.1K tokens at the first step to 32.6K at the 40th, and a median episode consumes 649K input tokens. Long histories also cause failures: in the main evaluation, 93 episodes (49 of them on \typeThree{}) ended because the request exceeded the provider's size limit, 69 for Claude models and 24 for GLM-5V-Turbo. We therefore test whether compressing the history can reduce token usage.

\paragraph{Framework}
We split the agent into three models (Figure~\ref{fig:prefilter-framework} in Appendix~\ref{app:agent-prior}). A \emph{Planner} (Gemini~3 Flash) decides at each round whether to answer or to keep exploring, in which case it issues a natural-language instruction. A \emph{Worker}, a fine-tuned Qwen2.5-VL-7B, turns the instruction into one atomic action given the current image (move forward, turn by a small or large angle, look up or down, zoom, or stop). A \emph{Compressor} (Gemini~2.5 Flash-Lite) reads the task, the most recent actions, and notes on the key frames stored in the trajectory memory, and writes a short \texttt{brief} of the exploration state and a \texttt{focus\_hint} that points to evidence relevant to the next decision. The Planner receives this summary in place of the raw textual history, together with the task and the key frames, so its textual context stays short while visual evidence remains available.

\paragraph{Token usage}
Compared with the previous harness without Compressor, our Planner--Worker--Compressor harness reduces Planner token usage from 11.06M to 4.36M on \typeOne{}, from 6.20M to 2.44M on \typeTwo{}, and from 11.18M to 4.80M on \typeThree{}---reductions of 60.6\%, 60.6\%, and 57.1\%, respectively (Table~\ref{tab:prefilter-ablation} in Appendix~\ref{app:agent-prior}). These counts cover only the Planner and exclude Worker and Compressor tokens.

\section{Conclusion}
\label{sec:conclusion}

\geobench{} tests whether VLM agents can combine world knowledge with active exploration in real Street View scenes. Across 23 agents, knowledge-heavy and exploration-heavy performance diverge, the best agent trails an expert reference, and many failures come from abstaining or exhausting the step budget without committing to an answer, rather than from misreading the scene; finding and counting objects remains hard for every agent. The environment's verifiable rewards and restorable state also make it a training ground for such agents, which we leave to future work.

\section*{Acknowledgment}
Sky Computing Lab is supported by gifts from Accenture, Algorithmic SuperIntelligence Labs, Amazon, AMD, Anyscale, Broadcom, cmpnd, Google, IBM, Intel, Intesa Sanpaolo, Lambda, Lightspeed, Mirendil, NVIDIA, Soma Labs, and VESSL.

\clearpage
\begingroup
\setlength{\bibsep}{1pt plus 0.5pt}
\bibliography{references}

\begin{thebibliography}{34}
\providecommand{\natexlab}[1]{#1}
\providecommand{\url}[1]{\texttt{#1}}
\expandafter\ifx\csname urlstyle\endcsname\relax
  \providecommand{\doi}[1]{doi: #1}\else
  \providecommand{\doi}{doi: \begingroup \urlstyle{rm}\Url}\fi

\bibitem[Yang et~al.(2023)Yang, Li, Wang, Lin, Azarnasab, Ahmed, Liu, Liu,
  Zeng, and Wang]{yang2023mmreact}
Zhengyuan Yang, Linjie Li, Jianfeng Wang, Kevin Lin, Ehsan Azarnasab, Faisal
  Ahmed, Zicheng Liu, Ce~Liu, Michael Zeng, and Lijuan Wang.
\newblock {MM-REACT}: Prompting {ChatGPT} for multimodal reasoning and action.
\newblock \emph{arXiv preprint arXiv:2303.11381}, 2023.

\bibitem[Liu et~al.(2023)Liu, Cheng, Liu, Zhang, Li, Ren, Zou, Yang, Su, Zhu,
  Zhang, Gao, and Li]{liu2023llavaplus}
Shilong Liu, Hao Cheng, Haotian Liu, Hao Zhang, Feng Li, Tianhe Ren, Xueyan
  Zou, Jianwei Yang, Hang Su, Jun Zhu, Lei Zhang, Jianfeng Gao, and Chunyuan
  Li.
\newblock {LLaVA-Plus}: Learning to use tools for creating multimodal agents.
\newblock \emph{arXiv preprint arXiv:2311.05437}, 2023.

\bibitem[Sur{\'\i}s et~al.(2023)Sur{\'\i}s, Menon, and
  Vondrick]{suris2023vipergpt}
D{\'\i}dac Sur{\'\i}s, Sachit Menon, and Carl Vondrick.
\newblock {ViperGPT}: Visual inference via python execution for reasoning.
\newblock In \emph{Proceedings of the IEEE/CVF International Conference on
  Computer Vision (ICCV)}, 2023.

\bibitem[Patil et~al.(2025)Patil, Mao, Yan, Ji, Suresh, Stoica, and
  Gonzalez]{patil2025bfcl}
Shishir~G. Patil, Huanzhi Mao, Fanjia Yan, Charlie Cheng-Jie Ji, Vishnu Suresh,
  Ion Stoica, and Joseph~E. Gonzalez.
\newblock The {Berkeley Function Calling Leaderboard (BFCL)}: From tool use to
  agentic evaluation of large language models.
\newblock In \emph{Proceedings of the 42nd International Conference on Machine
  Learning (ICML)}, volume 267 of \emph{Proceedings of Machine Learning
  Research}, pages 48371--48392, 2025.

\bibitem[Trivedi et~al.(2024)Trivedi, Khot, Hartmann, Manku, Dong, Li, Gupta,
  Sabharwal, and Balasubramanian]{trivedi2024appworld}
Harsh Trivedi, Tushar Khot, Mareike Hartmann, Ruskin Manku, Vinty Dong, Edward
  Li, Shashank Gupta, Ashish Sabharwal, and Niranjan Balasubramanian.
\newblock {AppWorld}: A controllable world of apps and people for benchmarking
  interactive coding agents.
\newblock In \emph{Proceedings of the 62nd Annual Meeting of the Association
  for Computational Linguistics (ACL)}, 2024.
\newblock arXiv:2407.18901.

\bibitem[Wang et~al.(2026)Wang, Zhang, Ge, Lian, Fu, Dunlap, Goldberg, Wang,
  Stoica, Chan, Min, and Gonzalez]{wang2026visgym}
Zirui Wang, Junyi Zhang, Jiaxin Ge, Long Lian, Letian Fu, Lisa Dunlap, Ken
  Goldberg, XuDong Wang, Ion Stoica, David~M. Chan, Sewon Min, and Joseph~E.
  Gonzalez.
\newblock {VisGym}: Diverse, customizable, scalable environments for multimodal
  agents.
\newblock \emph{arXiv preprint arXiv:2601.16973}, 2026.

\bibitem[Haas et~al.(2024)Haas, Skreta, Alberti, and Finn]{haas2024pigeon}
Lukas Haas, Michal Skreta, Silas Alberti, and Chelsea Finn.
\newblock Pigeon: Predicting image geolocations.
\newblock In \emph{Proceedings of the IEEE/CVF Conference on Computer Vision
  and Pattern Recognition}, pages 12893--12902, 2024.

\bibitem[{Plonk It}(n.d.)]{plonkit}
{Plonk It}.
\newblock Beginner's guide to {GeoGuessr}, n.d.
\newblock URL \url{https://www.plonkit.net/beginners-guide}.
\newblock Accessed 2026-03-20.

\bibitem[Zheng et~al.(2026)Zheng, Duan, Zhang, Liu, and Min]{zheng2026wander}
Yushuo Zheng, Huiyu Duan, Zicheng Zhang, Xiaohong Liu, and Xiongkuo Min.
\newblock Learning to wander: Improving the global image geolocation ability of
  {LMMs} via actionable reasoning.
\newblock \emph{arXiv preprint arXiv:2603.10463}, 2026.

\bibitem[Yi et~al.(2026)Yi, Mosk-Aoyama, Huang, Gala, Wang, Devare, Bhardwaj,
  Gupta, Kuchaiev, Jiao, Zhang, and Srinivasan]{yi2026pivotrl}
Junkeun Yi, Damon Mosk-Aoyama, Baihe Huang, Ritu Gala, Charles Wang,
  Sugam~Dipak Devare, Khushi Bhardwaj, Abhibha Gupta, Oleksii Kuchaiev, Jiantao
  Jiao, Jian Zhang, and Venkat Srinivasan.
\newblock {PivotRL}: High accuracy agentic post-training at low compute cost.
\newblock \emph{arXiv preprint arXiv:2603.21383}, 2026.

\bibitem[Weyand et~al.(2016)Weyand, Kostrikov, and Philbin]{weyand2016planet}
Tobias Weyand, Ilya Kostrikov, and James Philbin.
\newblock Planet - photo geolocation with convolutional neural networks.
\newblock \emph{CoRR}, abs/1602.05314, 2016.
\newblock URL \url{http://arxiv.org/abs/1602.05314}.

\bibitem[Theiner et~al.(2021)Theiner, M{\"{u}}ller{-}Budack, and
  Ewerth]{Theiner2021Interpretablesamanticphotogeolocalization}
Jonas Theiner, Eric M{\"{u}}ller{-}Budack, and Ralph Ewerth.
\newblock Interpretable semantic photo geolocalization.
\newblock \emph{CoRR}, abs/2104.14995, 2021.
\newblock URL \url{https://arxiv.org/abs/2104.14995}.

\bibitem[Pramanick et~al.(2022)Pramanick, Nowara, Gleason, Castillo, and
  Chellappa]{pramanick2022worldimagetransformerbasedgeolocalization}
Shraman Pramanick, Ewa~M. Nowara, Joshua Gleason, Carlos~D. Castillo, and Rama
  Chellappa.
\newblock Where in the world is this image? transformer-based geo-localization
  in the wild, 2022.
\newblock URL \url{https://arxiv.org/abs/2204.13861}.

\bibitem[Cepeda et~al.(2023)Cepeda, Nayak, and
  Shah]{cepeda2023geoclipclipinspiredalignmentlocations}
Vicente~Vivanco Cepeda, Gaurav~Kumar Nayak, and Mubarak Shah.
\newblock Geoclip: Clip-inspired alignment between locations and images for
  effective worldwide geo-localization, 2023.
\newblock URL \url{https://arxiv.org/abs/2309.16020}.

\bibitem[Zhou et~al.(2024)Zhou, Zhang, Guan, Hu, Lao, Mu, Li, and
  Mai]{zhou2024img2loc}
Zhongliang Zhou, Jielu Zhang, Zihan Guan, Mengxuan Hu, Ni~Lao, Lan Mu, Sheng
  Li, and Gengchen Mai.
\newblock Img2loc: Revisiting image geolocalization using multi-modality
  foundation models and image-based retrieval-augmented generation.
\newblock In \emph{Proceedings of the 47th International ACM SIGIR Conference
  on Research and Development in Information Retrieval}, SIGIR 2024, pages
  2749--2754. ACM, July 2024.
\newblock \doi{10.1145/3626772.3657673}.
\newblock URL \url{http://dx.doi.org/10.1145/3626772.3657673}.

\bibitem[Jia et~al.(2024)Jia, Liu, Li, Wang, Du, Han, Wei, Wang, Yin, and
  Zhao]{jia2024g3effectiveadaptiveframework}
Pengyue Jia, Yiding Liu, Xiaopeng Li, Yuhao Wang, Yantong Du, Xiao Han, Xuetao
  Wei, Shuaiqiang Wang, Dawei Yin, and Xiangyu Zhao.
\newblock G3: An effective and adaptive framework for worldwide geolocalization
  using large multi-modality models, 2024.
\newblock URL \url{https://arxiv.org/abs/2405.14702}.

\bibitem[Daruna et~al.(2026)Daruna, Meegan, Chiu, Samarasekera, and
  Kumar]{daruna2026geosurgegeolocalizationusingsemantic}
Angel Daruna, Nicholas Meegan, Han-Pang Chiu, Supun Samarasekera, and Rakesh
  Kumar.
\newblock Geosurge: Geo-localization using semantic fusion with hierarchy of
  geographic embeddings, 2026.
\newblock URL \url{https://arxiv.org/abs/2510.01448}.

\bibitem[Jia et~al.(2025)Jia, Park, Gao, Zhao, and
  Li]{jia2025georankerdistanceawarerankingworldwide}
Pengyue Jia, Seongheon Park, Song Gao, Xiangyu Zhao, and Sharon Li.
\newblock Georanker: Distance-aware ranking for worldwide image
  geolocalization, 2025.
\newblock URL \url{https://arxiv.org/abs/2505.13731}.

\bibitem[Li et~al.(2026)Li, Xiao, Wang, Ma, Yang, and
  Li]{li2026locationagenthierarchicalagentimage}
Qiujun Li, Zijin Xiao, Xulin Wang, Zhidan Ma, Cheng Yang, and Haifeng Li.
\newblock Locationagent: A hierarchical agent for image geolocation via
  decoupling strategy and evidence from parametric knowledge, 2026.
\newblock URL \url{https://arxiv.org/abs/2601.19155}.

\bibitem[Mirowski et~al.(2018)Mirowski, Grimes, Malinowski, Hermann, Anderson,
  Teplyashin, Simonyan, Kavukcuoglu, Zisserman, and
  Hadsell]{mirowski2018learning}
Piotr Mirowski, Matthew~Koichi Grimes, Mateusz Malinowski, Karl~Moritz Hermann,
  Keith Anderson, Denis Teplyashin, Karen Simonyan, Koray Kavukcuoglu, Andrew
  Zisserman, and Raia Hadsell.
\newblock Learning to navigate in cities without a map.
\newblock In \emph{Advances in Neural Information Processing Systems}, 2018.

\bibitem[Chen et~al.(2019)Chen, Suhr, Misra, Snavely, and
  Artzi]{chen2019touchdown}
Howard Chen, Alane Suhr, Dipendra Misra, Noah Snavely, and Yoav Artzi.
\newblock {TOUCHDOWN}: Natural language navigation and spatial reasoning in
  visual street environments.
\newblock In \emph{Proceedings of the IEEE/CVF Conference on Computer Vision
  and Pattern Recognition (CVPR)}, 2019.

\bibitem[Ashraf et~al.(2025)Ashraf, Saqib, Ghani, AlMahri, Li, Ahsan, Nawaz,
  Lahoud, Cholakkal, Shah, Torr, Khan, Anwer, and Khan]{ashraf2025agentx}
Tajamul Ashraf, Amal Saqib, Hanan Ghani, Muhra AlMahri, Yuhao Li, Noor Ahsan,
  Umair Nawaz, Jean Lahoud, Hisham Cholakkal, Mubarak Shah, Philip Torr,
  Fahad~Shahbaz Khan, Rao~Muhammad Anwer, and Salman Khan.
\newblock {Agent-X}: Evaluating deep multimodal reasoning in vision-centric
  agentic tasks.
\newblock \emph{arXiv preprint arXiv:2505.24876}, 2025.

\bibitem[Ning et~al.(2024)Ning, Su, Lv, Zhang, Liu, Liu, and
  Xu]{ning2024wtueval}
Kangyun Ning, Yisong Su, Xueqiang Lv, Yuanzhe Zhang, Jian Liu, Kang Liu, and
  Jinan Xu.
\newblock {WTU-EVAL}: A whether-or-not tool usage evaluation benchmark for
  large language models.
\newblock \emph{arXiv preprint arXiv:2407.12823}, 2024.

\bibitem[Ma et~al.(2024)Ma, Song, Zhuang, Hao, and King]{ma2024vla}
Yueen Ma, Zixing Song, Yuzheng Zhuang, Jianye Hao, and Irwin King.
\newblock A survey on vision-language-action models for embodied {AI}.
\newblock \emph{arXiv preprint arXiv:2405.14093}, 2024.

\bibitem[Yao et~al.(2023)Yao, Zhao, Yu, Du, Shafran, Narasimhan, and
  Cao]{yao2023react}
Shunyu Yao, Jeffrey Zhao, Dian Yu, Nan Du, Izhak Shafran, Karthik Narasimhan,
  and Yuan Cao.
\newblock {ReAct}: Synergizing reasoning and acting in language models.
\newblock In \emph{International Conference on Learning Representations
  (ICLR)}, 2023.
\newblock arXiv:2210.03629.

\bibitem[Schick et~al.(2023)Schick, Dwivedi-Yu, Dess{\`\i}, Raileanu, Lomeli,
  Zettlemoyer, Cancedda, and Scialom]{schick2023toolformer}
Timo Schick, Jane Dwivedi-Yu, Roberto Dess{\`\i}, Roberta Raileanu, Maria
  Lomeli, Luke Zettlemoyer, Nicola Cancedda, and Thomas Scialom.
\newblock Toolformer: Language models can teach themselves to use tools.
\newblock In \emph{Advances in Neural Information Processing Systems
  (NeurIPS)}, 2023.
\newblock arXiv:2302.04761.

\bibitem[Patil et~al.(2023)Patil, Zhang, Wang, and Gonzalez]{patil2023gorilla}
Shishir~G. Patil, Tianjun Zhang, Xin Wang, and Joseph~E. Gonzalez.
\newblock Gorilla: Large language model connected with massive {API}s.
\newblock \emph{arXiv preprint arXiv:2305.15334}, 2023.

\bibitem[Qin et~al.(2023)Qin, Hu, Lin, Chen, Ding, Cui, Zeng, Huang, Xiao, Han,
  Fung, Su, Wang, Qian, Tian, Zhu, Liang, Shen, Xu, Zhang, Ye, Li, Tang, Yi,
  Zhu, Dai, Yan, Cong, Lu, Zhao, Huang, Yan, Han, Sun, Li, Phang, Yang, Wu, Ji,
  Liu, and Sun]{qin2023tool}
Yujia Qin, Shengding Hu, Yankai Lin, Weize Chen, Ning Ding, Ganqu Cui, Zheni
  Zeng, Yufei Huang, Chaojun Xiao, Chi Han, Yi~Ren Fung, Yusheng Su, Huadong
  Wang, Cheng Qian, Runchu Tian, Kunlun Zhu, Shihao Liang, Xingyu Shen, Bokai
  Xu, Zhen Zhang, Yining Ye, Bowen Li, Ziwei Tang, Jing Yi, Yuzhang Zhu,
  Zhenning Dai, Lan Yan, Xin Cong, Yaxi Lu, Weilin Zhao, Yuxiang Huang, Junxi
  Yan, Xu~Han, Xian Sun, Dahai Li, Jason Phang, Cheng Yang, Tongshuang Wu, Heng
  Ji, Zhiyuan Liu, and Maosong Sun.
\newblock Tool learning with foundation models.
\newblock \emph{arXiv preprint arXiv:2304.08354}, 2023.

\bibitem[Dede(n.d.)]{Latb}
Nicolas Dede.
\newblock The maths of {GeoGuessr}, n.d.
\newblock URL \url{https://latb.io/geoguessr/articles/the-maths}.
\newblock Accessed 2026-04-23.

\bibitem[Sitzmann et~al.(2018)Sitzmann, Serrano, Pavel, Agrawala, Gutierrez,
  Masia, and Wetzstein]{sitzmann2018saliency}
Vincent Sitzmann, Ana Serrano, Amy Pavel, Maneesh Agrawala, Diego Gutierrez,
  Belen Masia, and Gordon Wetzstein.
\newblock Saliency in vr: How do people explore virtual environments?
\newblock \emph{IEEE Transactions on Visualization and Computer Graphics},
  24\penalty0 (4):\penalty0 1633--1642, 2018.
\newblock \doi{10.1109/TVCG.2018.2793599}.

\bibitem[Shinn et~al.(2023)Shinn, Cassano, Berman, Gopinath, Narasimhan, and
  Yao]{shinn2023reflexion}
Noah Shinn, Federico Cassano, Edward Berman, Ashwin Gopinath, Karthik
  Narasimhan, and Shunyu Yao.
\newblock Reflexion: Language agents with verbal reinforcement learning.
\newblock In \emph{Advances in Neural Information Processing Systems
  (NeurIPS)}, 2023.
\newblock arXiv:2303.11366.

\bibitem[{Significant Gravitas}(2023)]{autogpt2023}
{Significant Gravitas}.
\newblock {Auto-GPT}: An autonomous {GPT}-4 experiment.
\newblock \url{https://github.com/Significant-Gravitas/AutoGPT}, 2023.

\bibitem[Wang et~al.(2023)Wang, Xie, Jiang, Mandlekar, Xiao, Zhu, Fan, and
  Anandkumar]{wang2023voyager}
Guanzhi Wang, Yuqi Xie, Yunfan Jiang, Ajay Mandlekar, Chaowei Xiao, Yuke Zhu,
  Linxi Fan, and Anima Anandkumar.
\newblock Voyager: An open-ended embodied agent with large language models.
\newblock \emph{arXiv preprint arXiv:2305.16291}, 2023.

\bibitem[Snyder(1987)]{snyder1987map}
John~P. Snyder.
\newblock \emph{Map Projections: A Working Manual}.
\newblock Number 1395 in U.S. Geological Survey Professional Paper. U.S.
  Government Printing Office, Washington, D.C., 1987.
\newblock \doi{10.3133/pp1395}.

\end{thebibliography}
\endgroup

\clearpage
\beginsupplement
\crefalias{section}{appendix}
\raggedbottom
\makeatletter
\setlength{\@fptop}{0pt}
\setlength{\@fpsep}{16pt}
\setlength{\@fpbot}{0pt plus 1fil}
\makeatother

\section*{Appendix Contents}
\phantomsection
\label{sec:appendix}

\newcommand{\appendixentry}[2]{%
  \noindent\hspace*{#1}\hyperref[#2]{\ref*{#2}\quad\nameref*{#2}}%
  \nobreak\dotfill\hyperref[#2]{\pageref*{#2}}\par}
\begingroup
\small
\setlength{\parskip}{2pt}
\hypersetup{linkcolor=black}
{\bfseries\appendixentry{0pt}{app:agent-prior}}
\smallskip
{\bfseries\appendixentry{0pt}{app:type1-scoring}}
\smallskip
{\bfseries\appendixentry{0pt}{app:answer-matching}}
\smallskip
{\bfseries\appendixentry{0pt}{app:visual-context}}
\smallskip
{\bfseries\appendixentry{0pt}{app:fm-annotation}}
\smallskip
{\bfseries\appendixentry{0pt}{app:additional}}
\endgroup

\section{Planner--Worker--Compressor Details}
\label{app:agent-prior}

\begin{figure}[!htbp]
\centering
\includegraphics[width=\columnwidth]{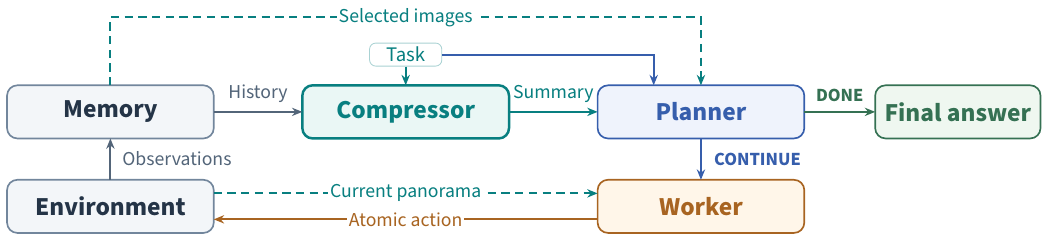}
\caption{\textbf{Planner--Worker--Compressor framework.} The Compressor condenses recent actions and key-frame notes into a \texttt{brief} and \texttt{focus\_hint}; selected images remain available to the Planner. The Planner either returns an answer or instructs the Worker to act, producing new observations for Memory. Dashed arrows denote image inputs to the models.}
\label{fig:prefilter-framework}
\end{figure}

\begin{table}[!htbp]
\centering
\small
\setlength{\tabcolsep}{4pt}
\caption{\textbf{Planner tokens and task results with and without Compressor.} M denotes millions of tokens; reductions are computed from the displayed counts. \typeOne{} reports the GeoGuessr score (the mean distance error falls from 166.89~km to 89.56~km with Compressor); \typeTwo{} and \typeThree{} report accuracy.}
\label{tab:prefilter-ablation}
\begin{tabular}{lccc}
\toprule
 & \taskNameCell{\typeOne} & \taskNameCell{\typeTwo} & \taskNameCell{\typeThree} \\
\midrule
Planner tokens, no Compressor & 11.06M & 6.20M & 11.18M \\
Planner tokens, Compressor    & 4.36M  & 2.44M & 4.80M  \\
Token reduction              & 60.6\% & 60.6\% & 57.1\% \\
\midrule
Result, no Compressor         & 4686   & 30.00\% & 24.39\% \\
Result, Compressor            & 4737   & 26.25\% & 24.39\% \\
\bottomrule
\end{tabular}
\end{table}

\paragraph{Configuration} The Planner is \texttt{gemini-\allowbreak 3-\allowbreak flash-\allowbreak preview} and the Compressor \texttt{gemini-\allowbreak 2.5-\allowbreak flash-\allowbreak lite}. The Planner's output has four fields, \texttt{thought}, \texttt{status}, \texttt{instruction}, and \texttt{answer}: status \texttt{CONTINUE} passes the instruction to the Worker, and \texttt{DONE} returns the answer. Table~\ref{tab:prefilter-ablation} reports planner tokens and task results with and without Compressor.

\paragraph{Relation to prior agent frameworks}
ReAct~\citep{yao2023react} interleaves chain-of-thought reasoning with tool actions in a closed reason--act--observe loop. Reflexion~\citep{shinn2023reflexion} adds verbal self-criticism to this loop; Auto-GPT~\citep{autogpt2023} lets a single LLM decompose a goal and dispatch sub-tasks to itself; and Voyager~\citep{wang2023voyager} accumulates a library of executable skills through interaction in an embodied setting. These frameworks share the premise that long-horizon problems are best solved by alternating reasoning with observation. The framework in Section~\ref{sec:method} follows this premise but departs from prior work in two ways: it splits high-level reasoning and low-level visual action across two models of very different capacity, and it places an explicit Compressor between trajectory memory and the planner, which separates context summarization from high-level reasoning.

\section{\typeOne{} Scoring}
\label{app:type1-scoring}
\begin{figure}[htbp]
    \centering
    \includegraphics[width=\linewidth]{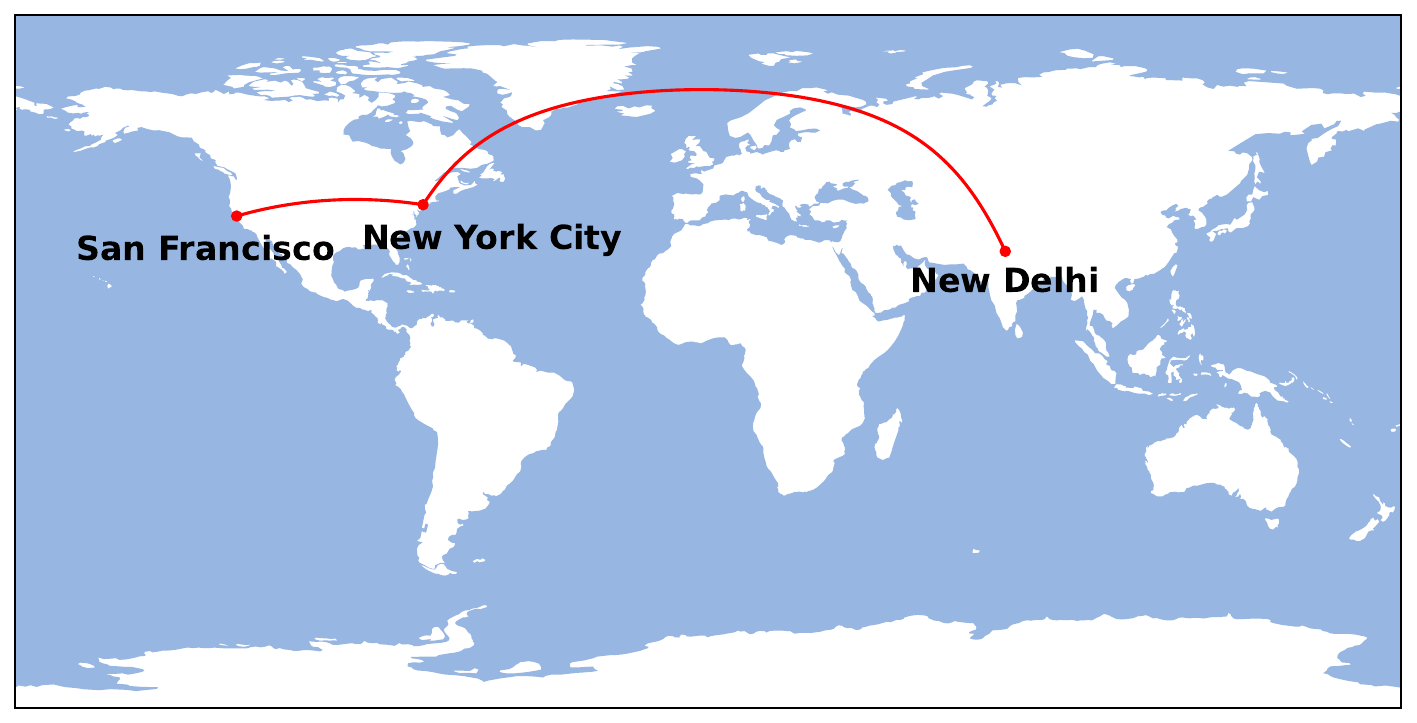}
    \caption{\textbf{Haversine Distance Visualization.} Haversine distance from New York City to San Francisco (\textasciitilde4129 km; score=313) and New Delhi (\textasciitilde11755 km; score=2).}
    \label{fig:formula_map}
\end{figure}

The score $S(d)$ (Section~\ref{sec:eval_setup}) halves with every $D\ln 2/10 \approx 1{,}034$~km of error (Figure~\ref{fig:formula_map}). We average scores rather than distances because the score is defined for invalid responses, which receive 0, and is not dominated by a few intercontinental misses. Tables also report the score-equivalent distance
\begin{equation}
    \bar d = -\frac{D}{10}\ln\frac{\bar S}{5000},
    \label{eq:score-equivalent-distance}
\end{equation}
the error whose score equals the mean score $\bar S$; it is not the mean error.

\paragraph{Distance computation}
GeoGuessr measures guess error with the haversine distance~\citep{Latb}. Given two latitude-longitude coordinates $(\phi_1, \lambda_1)$ and $(\phi_2, \lambda_2)$ on a spherical surface of radius $r$, the equation for their haversine distance is given by~\citep{snyder1987map}:

\begin{equation}
\text{Hav}(\mathbf{x_1,x_2}) =
2r\arcsin{\left(
    \sqrt{
        \sin^2{
                \left( \frac{\phi_2-\phi_1}{2} \right)
            }
        +
        \cos{(\phi_1)}\cos{(\phi_2)}
        \sin^2{
                \left( \frac{\lambda_2-\lambda_1}{2} \right)
            }
    }
\right)}
\end{equation}

This quantity represents the shortest distance between the two coordinates over the surface of the sphere; GeoGuessr uses $r = 6371$~km~\citep{Latb}. Our evaluator instead computes the geodesic distance on the WGS-84 ellipsoid with the \texttt{geopy} library. For any pair of points, the two distances differ by less than 0.6\%, which changes the score $S(d)$ (Section~\ref{sec:eval_setup}) by at most about 10 points.

\section{\typeTwo{} and \typeThree{} Scoring}
\label{app:answer-matching}
We extract the \texttt{answer} field of the final reply, lowercase it and each accepted answer, and strip punctuation; the reply is correct only if the result equals one of the entry's accepted answers. Strict matching gives no credit for hedging: an answer that lists several candidates (``Kenya or Uganda'') or negates one (``not Kenya'') is incorrect, whereas a looser rule such as whole-word containment would accept both when ``Kenya'' is accepted. Its risk is the opposite error, a correct answer written in a form that the accepted answers do not list; to limit it, the accepted answers cover equivalent forms of each reference answer, such as ``Czechia'' and ``Czech Republic'' (Section~\ref{subsec:curation}).

\paragraph{Label audit}
We audited the automatic labels for both kinds of error. For false positives, we checked replies marked correct for answers that are hedged, negated, or otherwise wrong; for false negatives, we checked replies marked incorrect for answers equivalent to the reference answer. The audit found no mislabeled reply.

\section{Visual Context Measurement}
\label{app:visual-context}
\begin{figure}[t]
    \centering
    \includegraphics[width=\linewidth]{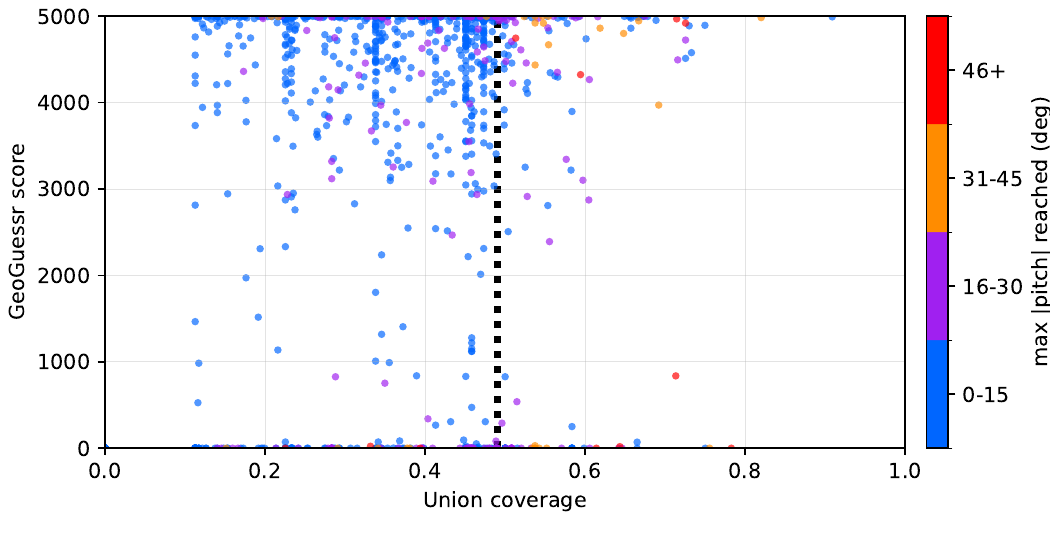}
    \caption{\textbf{Panorama coverage versus performance--No Movement, \typeOne{}.} Observations from ten models on Competition and Synthetic are pooled in one plot of panorama coverage (Eq.~\ref{eq:visual-coverage}) against GeoGuessr score; color denotes maximum absolute pitch. Invalid responses remain at zero score, and zero coverage denotes an empty reconstructed trace. The dotted line at approximately 0.49 indicates the coverage limit of a full horizon sweep at the default field of view; pitch changes or wider fields of view can exceed it.}
    \label{fig:coverage-performance}
\end{figure}

The no-movement coverage analysis uses ten of the 13 ablation models (Figure~\ref{fig:coverage-performance}); Claude Opus 4.6, GPT-5.2, and Qwen3.5-397B are excluded because their camera traces were not available for this analysis. Each model contributes \numCompetition{} \typeCompetition{} and \numTypeOne{} Synthetic episodes, including zero-score responses.

\paragraph{Coverage definition}
For an episode confined to one panorama, angular coverage is the fraction of the viewing sphere covered by the union of reconstructed camera views:
\begin{equation}
    C = \frac{1}{4\pi}\int_{\mathbb{S}^{2}} f(\mathbf{d})\,d\Omega,
    \qquad
    f(\mathbf{d}) =
    \begin{cases}
        1, & \text{if } \mathbf{d}\in\bigcup_t F_t,\\
        0, & \text{otherwise}.
    \end{cases}
    \label{eq:visual-coverage}
\end{equation}
where $\mathbf{d}$ is a unit viewing direction and $F_t$ is the spherical footprint of the perspective view after tool call $t$. Overlapping views count once. Figure~\ref{fig:nomovement_figure} illustrates this union in an equirectangular projection, whose areas require latitude weighting to measure spherical coverage. Coverage measures reconstructed view availability, not semantic information or the agent's attention: zooming into a sign may improve legibility without adding angular area.

\paragraph{Camera replay and spherical integration}
We reconstruct camera states by replaying each trace from heading, pitch, and horizontal field of view $(0^\circ,0^\circ,90^\circ)$. The replay updates heading and pitch with each \texttt{scroll\_*} call, and divides or multiplies the field of view by the zoom factor for \texttt{zoom\_in} or \texttt{zoom\_out}, respectively. Pitch is clamped to $\pm90^\circ$ and field of view to $[10^\circ,120^\circ]$. Each recorded tool call emits one reconstructed view, including multiple calls within an agentic step. No extra initial view is inserted. An empty trace is assigned zero coverage, meaning no reconstructed tool-call views, rather than proof that the agent received no initial visual input. This replay assumes that recorded camera commands took effect; it does not independently validate tool execution or image delivery.

For each view, a unit direction belongs to its footprint if it lies in front of the camera and its horizontal and vertical perspective coordinates fall within the field-of-view bounds. We form the Boolean union of these footprints on a $2048\times1024$ longitude--latitude grid. A cell centered at latitude $\phi$ contributes $\cos\phi\,\Delta\phi\,\Delta\lambda$ steradians; division by $4\pi$ gives Eq.~\ref{eq:visual-coverage}. The raster grid controls numerical integration resolution and is distinct from the $1920\times1080$ rendered viewport. As a geometric check, a single $90^\circ$ horizontal view at this aspect ratio covers approximately 0.1127 of the sphere.

\begin{figure}[!htbp]
    \centering
    \includegraphics[width=0.55\linewidth]{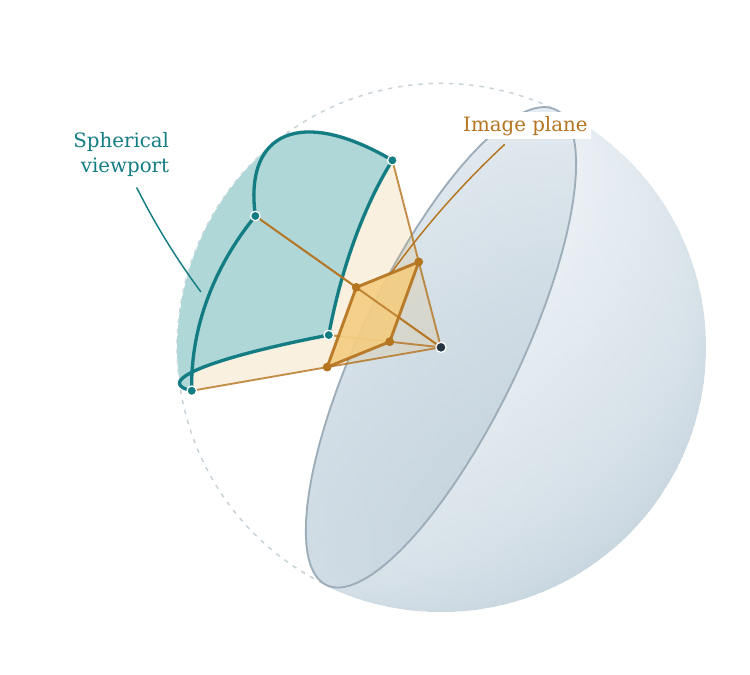}
    \caption{\textbf{Camera field of view on the unit sphere.} A perspective camera at the origin projects rays through a rectangular image plane (gold) onto the unit sphere. Normalizing each ray yields the spherical viewport (teal), the footprint $F_t$ of a single view in Eq.~\ref{eq:visual-coverage}, whose curved boundary is set by the camera field of view rather than by a latitude--longitude rectangle. The default $h=90^\circ$ horizontal field of view and $16{:}9$ aspect ratio give $v\approx58.72^\circ$ vertically. The cutaway removes the forward hemisphere to expose the observer and viewing pyramid; the dashed silhouette marks the full sphere.}
    \label{fig:camera-fov-sphere}
\end{figure}

\paragraph{Horizontal-sweep coverage}
\label{app:horizontal-sweep}
Figure~\ref{fig:camera-fov-sphere} shows how a single perspective view maps onto the sphere. For the $1920\times1080$ viewport and default horizontal field of view $h=90^\circ$, the vertical field of view is
$v=2\arctan[(1080/1920)\tan(h/2)]\approx58.72^\circ$.
A dense horizontal sweep at zero pitch therefore covers a belt with fraction $\sin(v/2)\approx0.4903$ of the sphere. This corresponds to the vertical cutoff in coverage values pooled across models and datasets in Figure~\ref{fig:coverage-performance}. The 0.4903 bound applies to horizon-centered views with horizontal field of view at most $90^\circ$; either changing pitch or zooming out beyond $90^\circ$ can exceed it.

\paragraph{Coverage and performance}
Figure~\ref{fig:coverage-performance} plots angular coverage against GeoGuessr score, pooling observations across ten models and the Competition and Synthetic subsets. Among traces with at least one tool call, 254/380 (66.8\%) on Competition and 496/738 (67.2\%) on Synthetic keep pitch at zero throughout. The remaining 40 and 62 entries, respectively, have no recorded tool calls; they remain in the plot at $C=0$ but are excluded from these behavioral percentages. This concentration at the horizon limits angular coverage even when agents turn through all headings. Color denotes the maximum absolute pitch reached, with $0^\circ$--$15^\circ$ grouped in the same bin; exact zero-pitch counts are computed from the traces, not inferred from color. Discrete rotation choices produce repeated coverage values, and the plot is descriptive rather than a controlled dose--response experiment.

High scores do not require broad angular coverage. For example, Opus 4.8 never changes pitch in its 42 \typeCompetition{} no-movement traces, has a median coverage of 0.3075, and achieves a mean score of 4646. GLM-5V-Turbo has a greater median coverage, 0.4732, but a lower mean score of 4249 on the same split. This comparison illustrates why angular extent alone cannot explain performance: models also differ in which regions they inspect and how they interpret them. It does not establish that narrower coverage improves accuracy. Figure~\ref{fig:nomovement_figure} illustrates the distinction between inspecting a panorama and exposing its entire surface.

\section{Failure-Mode Annotation}
\label{app:fm-annotation}
Five annotators labeled the 865 failed episodes of the four models in Section~\ref{subsec:failure_modes}. The episodes were divided among the annotators, and each annotator inspected the traces assigned to them and gave each episode one or more of the three failure modes in Table~\ref{tab:failure-modes}. To measure agreement, 300 episodes were labeled by more than one annotator, and the annotators' labels agreed on 95\% of them.

\clearpage
\section{Additional Figures and Tables}
\label{app:additional}

\begin{table*}[h]
  \centering
  \small
  \setlength{\tabcolsep}{3pt}
  \caption{\textbf{Average steps per episode (mean $\pm$ standard deviation).} Number of agent steps (tool-call rounds plus the final answer) the model takes before committing a response, per task split. Statistics are computed over \emph{completed} episodes only; episodes that ended in an API error, such as requests exceeding the provider's size limit (HTTP 413) on long, image-heavy episodes of Claude Fable 5 and Claude Opus 4.6, are excluded. Claude Opus rows use the direct Anthropic endpoint. The step budget is 40.}
  \label{tab:step-counts}
  \begin{tabular}{lcccc}
    \toprule
    & \multicolumn{2}{c}{\textbf{\typeOne}} & \textbf{\typeTwo} & \textbf{\typeThree} \\
    \cmidrule(lr){2-3}
    \textbf{Model} &
    \textbf{\makecell[b]{\typeCompetition \\ Steps}} &
    \textbf{\makecell[b]{Synthetic \\ Steps}} &
    \textbf{Steps} &
    \textbf{Steps} \\
    \midrule
  Claude Fable 5       & 29.1 $\pm$ 12.5 & 23.9 $\pm$ 15.3 & 20.0 $\pm$ 16.0 & 31.1 $\pm$ 11.1 \\
  Claude Opus 4.8      & 12.8 $\pm$ 8.9  & 9.7 $\pm$ 7.9   & 11.6 $\pm$ 11.3 & 27.0 $\pm$ 13.2 \\
  Claude Opus 4.7      & 11.8 $\pm$ 8.0  & 9.8 $\pm$ 9.0   & 14.7 $\pm$ 13.4 & 25.1 $\pm$ 12.3 \\
  Claude Opus 4.6      & 24.4 $\pm$ 12.5 & 24.5 $\pm$ 14.5 & 24.8 $\pm$ 15.3 & 33.5 $\pm$ 10.7 \\
  Claude Sonnet 4.6    & 29.4 $\pm$ 13.1 & 23.2 $\pm$ 13.9 & 24.1 $\pm$ 15.3 & 34.5 $\pm$ 10.3 \\
  GPT-5.5 Pro          & 38.3 $\pm$ 6.0  & 36.0 $\pm$ 8.3  & 24.6 $\pm$ 15.6 & 30.0 $\pm$ 11.4 \\
  GPT-5.5              & 30.0 $\pm$ 8.8  & 27.2 $\pm$ 11.1 & 17.4 $\pm$ 12.2 & 24.9 $\pm$ 10.2 \\
  GPT-5.4 Pro          & 35.7 $\pm$ 7.2  & 33.2 $\pm$ 9.3  & 23.4 $\pm$ 12.2 & 25.8 $\pm$ 10.3 \\
  GPT-5.4              & 10.8 $\pm$ 2.7  & 10.1 $\pm$ 3.0  & 7.2 $\pm$ 3.5   & 10.7 $\pm$ 8.1  \\
  GPT-5.2              & 9.9 $\pm$ 2.9   & 9.3 $\pm$ 3.1   & 9.1 $\pm$ 5.5   & 12.6 $\pm$ 9.4  \\
  Gemini 3.1 Pro       & 39.5 $\pm$ 5.4  & 37.0 $\pm$ 8.8  & 29.1 $\pm$ 14.0 & 32.8 $\pm$ 12.0 \\
  Gemini 3.5 Flash     & 35.2 $\pm$ 9.2  & 33.7 $\pm$ 10.8 & 30.1 $\pm$ 13.7 & 35.7 $\pm$ 10.1 \\
  Gemini 3 Flash       & 28.3 $\pm$ 13.1 & 28.7 $\pm$ 14.0 & 31.4 $\pm$ 13.2 & 36.4 $\pm$ 8.9  \\
  Kimi K2.6            & 26.3 $\pm$ 12.1 & 22.3 $\pm$ 13.2 & 24.3 $\pm$ 13.2 & 31.9 $\pm$ 12.1 \\
  Kimi K2.5            & 20.9 $\pm$ 10.5 & 21.8 $\pm$ 10.8 & 20.6 $\pm$ 12.5 & 32.9 $\pm$ 11.1 \\
  Grok 4.3             & 3.3 $\pm$ 1.4   & 3.5 $\pm$ 1.3   & 3.5 $\pm$ 1.6   & 4.1 $\pm$ 3.0   \\
  GLM-5V-Turbo         & 22.0 $\pm$ 11.0 & 18.5 $\pm$ 9.0  & 16.4 $\pm$ 12.0 & 22.3 $\pm$ 13.3 \\
  MiniMax-M3           & 13.9 $\pm$ 11.3 & 11.6 $\pm$ 8.6  & 7.8 $\pm$ 9.7   & 25.2 $\pm$ 14.9 \\
  Qwen3.7-Plus         & 12.6 $\pm$ 4.3  & 11.2 $\pm$ 5.2  & 9.8 $\pm$ 8.8   & 20.3 $\pm$ 13.8 \\
  Qwen3.6-Plus         & 4.0 $\pm$ 3.6   & 4.1 $\pm$ 3.8   & 3.6 $\pm$ 3.1   & 16.6 $\pm$ 14.4 \\
  Qwen3.5-397B-A17B    & 12.0 $\pm$ 7.6  & 15.4 $\pm$ 8.0  & 8.1 $\pm$ 8.2   & 16.7 $\pm$ 12.9 \\
  Qwen3.5-122B-A10B    & 3.5 $\pm$ 2.0   & 3.6 $\pm$ 2.0   & 2.5 $\pm$ 1.1   & 9.6 $\pm$ 10.5  \\
  Qwen3.5-35B-A3B      & 4.2 $\pm$ 2.1   & 4.5 $\pm$ 2.4   & 6.2 $\pm$ 5.6   & 8.8 $\pm$ 7.6   \\
    \bottomrule
  \end{tabular}
\end{table*}

\begin{table}[ht]
  \centering
  \small
  \setlength{\tabcolsep}{5pt}
  \caption{\textbf{Average movement calls per episode.} Mean number of \texttt{move\_*} tool calls per episode for the four strongest models by mean \typeOne{} score (\typeCompetition{} and Synthetic), by task type; plotted in Figure~\ref{fig:exploration-granularity}a. Counts are over completed episodes only.}
  \label{tab:movement-calls}
    \begin{tabular}{lcccc}
    \toprule
    \textbf{Model} & \textbf{Comp.} & \textbf{Synth.} & \textbf{\taskNameCell{\typeTwo}} & \textbf{\taskNameCell{\typeThree}} \\
    \midrule
  Claude Opus 4.8  & 3.07  & 2.33  & 3.15 & 8.68 \\
  Claude Opus 4.7  & 2.43  & 2.34  & 7.44 & 9.28 \\
  GPT-5.4          & 2.83  & 2.48  & 1.74 & 3.30 \\
  Qwen3.7-Plus     & 2.90  & 2.58  & 2.17 & 8.17 \\
    \bottomrule
  \end{tabular}
\end{table}

\begin{table}[ht]
  \centering
  \small
  \setlength{\tabcolsep}{5pt}
  \caption{\textbf{\typeTwo{} accuracy by question granularity.} Accuracy (\%) of the five strongest models on \typeTwo{}, by the geographic granularity the question asks for; plotted in Figure~\ref{fig:exploration-granularity}b. \textbf{Bold}: best per granularity.}
  \label{tab:type2-granularity}
    \begin{tabular}{lcccc}
    \toprule
    \textbf{Model} & \makecell{\textbf{Country}\\($n{=}27$)} & \makecell{\textbf{State}\\($n{=}11$)} & \makecell{\textbf{City}\\($n{=}31$)} & \makecell{\textbf{Street}\\($n{=}11$)} \\
    \midrule
  Claude Fable 5   & \textbf{88.9} & 27.3          & \textbf{58.1} & \textbf{45.5} \\
  Claude Opus 4.8  & \textbf{88.9} & \textbf{72.7} & 41.9          & 27.3          \\
  Claude Opus 4.7  & 81.5          & 54.5          & 45.2          & 27.3          \\
  GPT-5.5          & 85.2          & \textbf{72.7} & 48.4          & 36.4          \\
  GPT-5.4 Pro      & 81.5          & 54.5          & 51.6          & 36.4          \\
    \bottomrule
  \end{tabular}
\end{table}

\begin{table*}[t]
  \centering
  \small
  \setlength{\tabcolsep}{3pt}
  \caption{\textbf{\typeOne{} score distribution and failure causes.} For each \typeOne{} split: the median GeoGuessr score, the number of zero-score episodes (\typeCompetition{}: of \numCompetition; Synthetic: of \numTypeOne), and how many of them ended at the 40-step limit, with no answer in the final turn (Step; Section~\ref{sec:eval_setup}), or with an abstention before the limit (Abst.); the remaining zeros are API errors or episodes without a final reply. The last two columns count the \typeTwo{} and \typeThree{} episodes (of 80) that ended at the step limit in the same sense. Rows follow Table~\ref{tab:results-summary}.}
  \label{tab:zero-causes}
  \begin{tabular*}{\textwidth}{@{\extracolsep{\fill}}lcccccccccc@{}}
    \toprule
    & \multicolumn{4}{c}{\textbf{\typeCompetition}} & \multicolumn{4}{c}{\textbf{Synthetic}} & \multicolumn{2}{c}{\textbf{Step limit}} \\
    \cmidrule(lr){2-5} \cmidrule(lr){6-9} \cmidrule(lr){10-11}
    \textbf{Model} & Median & Zero & Step & Abst. & Median & Zero & Step & Abst. & \taskNameCell[50pt]{\typeTwo} & \taskNameCell[50pt]{\typeThree} \\
    \midrule
  Claude Opus 4.8    & 4923 & 2 & 2 & 0 & 4984 & 2 & 2 & 0 & 4 & 27 \\
  Claude Opus 4.7    & 4949 & 4 & 1 & 3 & 4988 & 4 & 2 & 2 & 9 & 17 \\
  GPT-5.5            & 4763 & 7 & 7 & 0 & 4968 & 18 & 18 & 0 & 7 & 10 \\
  Claude Fable 5     & 4872 & 18 & 15 & 0 & 4994 & 29 & 28 & 0 & 22 & 28 \\
  Qwen3.7-Plus       & 4411 & 2 & 0 & 1 & 4608 & 1 & 0 & 1 & 3 & 17 \\
  GPT-5.4 Pro        & 3662 & 20 & 20 & 0 & 3988 & 35 & 35 & 0 & 11 & 10 \\
  GPT-5.4            & 4209 & 1 & 0 & 1 & 4871 & 0 & 0 & 0 & 0 & 1 \\
  GLM-5V-Turbo       & 4091 & 11 & 6 & 0 & 4892 & 8 & 3 & 0 & 7 & 16 \\
  Claude Opus 4.6    & 4416 & 14 & 9 & 0 & 4978 & 22 & 21 & 0 & 24 & 37 \\
  MiniMax-M3         & 4138 & 12 & 4 & 6 & 4119 & 23 & 1 & 11 & 3 & 24 \\
  Qwen3.6-Plus       & 3691 & 6 & 0 & 6 & 3481 & 6 & 0 & 3 & 0 & 13 \\
  GPT-5.5 Pro        & 0 & 32 & 32 & 0 & 0 & 53 & 53 & 0 & 27 & 30 \\
  Gemini 3 Flash     & 4400 & 17 & 17 & 0 & 2420 & 39 & 39 & 0 & 47 & 56 \\
  Claude Sonnet 4.6  & 4460 & 19 & 15 & 0 & 4841 & 29 & 23 & 0 & 27 & 39 \\
  Gemini 3.5 Flash   & 0 & 27 & 27 & 0 & 0 & 44 & 44 & 0 & 35 & 53 \\
  Gemini 3.1 Pro     & 0 & 37 & 35 & 2 & 0 & 62 & 62 & 0 & 38 & 48 \\
  Kimi K2.5          & 0 & 23 & 1 & 22 & 4123 & 30 & 5 & 25 & 13 & 39 \\
  Kimi K2.6          & 0 & 27 & 8 & 19 & 0 & 43 & 14 & 29 & 18 & 41 \\
  Qwen3.5-397B-A17B  & 1954 & 14 & 1 & 13 & 3000 & 25 & 1 & 23 & 1 & 10 \\
  GPT-5.2            & 0 & 37 & 0 & 37 & 0 & 57 & 0 & 57 & 0 & 2 \\
  Grok 4.3           & 0 & 34 & 0 & 34 & 0 & 52 & 0 & 52 & 0 & 0 \\
  Qwen3.5-122B-A10B  & 0 & 40 & 0 & 39 & 0 & 62 & 0 & 61 & 0 & 5 \\
  Qwen3.5-35B-A3B    & 0 & 37 & 0 & 1 & 0 & 66 & 0 & 5 & 0 & 1 \\
    \bottomrule
  \end{tabular*}
\end{table*}

\begin{table*}[t]
  \centering
  \small
  \setlength{\tabcolsep}{3pt}
  \caption{\textbf{Score on answered episodes.} Mean GeoGuessr score\,$|$\,distance in km (Eq.~\ref{eq:score-equivalent-distance}) over episodes with a non-zero score on the two \typeOne{} splits, \typeCompetition{} ($n{=}\numCompetition$) and Synthetic ($n{=}\numTypeOne$), and the percentage of episodes that score zero, pooled over both ($n{=}122$). Zero scores come from abstentions, episodes that end at the step limit, and errors (Table~\ref{tab:zero-causes}). \textbf{Bold} marks the highest score and the lowest zero-score rate.}
  \label{tab:valid-answers}
  \begin{tabular*}{\textwidth}{@{\extracolsep{\fill}}lccc@{}}
    \toprule
    \textbf{Model} &
    \textbf{\makecell{\typeCompetition \\ Score\,$|$\,Dist}} &
    \textbf{\makecell{Synthetic \\ Score\,$|$\,Dist}} &
    \textbf{\makecell{Zero-score \\ rate \%}} \\
    \midrule
  Claude Opus 4.8      & 4570.0\,$|$\,134.1 & 4666.2\,$|$\,103.1 & 3.28 \\
  Claude Opus 4.7      & 4759.7\,$|$\,73.5 & 4738.3\,$|$\,80.2 & 6.56 \\
  GPT-5.5              & 4578.1\,$|$\,131.5 & 4474.6\,$|$\,165.6 & 20.49 \\
  Claude Fable 5       & 4937.0\,$|$\,18.9 & 4963.4\,$|$\,11.0 & 38.52 \\
  Qwen3.7-Plus         & 4162.6\,$|$\,273.4 & 4046.5\,$|$\,315.6 & 2.46 \\
  GPT-5.4 Pro          & 4780.0\,$|$\,67.1 & 4742.0\,$|$\,79.0 & 45.08 \\
  GPT-5.4              & 4018.8\,$|$\,325.9 & 4221.1\,$|$\,252.6 & \textbf{0.82} \\
  GLM-5V-Turbo         & 4251.8\,$|$\,241.8 & 4288.7\,$|$\,228.9 & 15.57 \\
  Claude Opus 4.6      & 4535.9\,$|$\,145.3 & 4639.3\,$|$\,111.7 & 29.51 \\
  MiniMax-M3           & 4286.2\,$|$\,229.8 & 4078.3\,$|$\,303.9 & 28.69 \\
  Qwen3.6-Plus         & 3553.1\,$|$\,509.6 & 3196.1\,$|$\,667.5 & 9.84 \\
  GPT-5.5 Pro          & 4886.3\,$|$\,34.3 & 4610.7\,$|$\,120.9 & 69.67 \\
  Gemini 3 Flash       & 4789.6\,$|$\,64.1 & 4864.9\,$|$\,40.9 & 45.90 \\
  Claude Sonnet 4.6    & 4837.5\,$|$\,49.3 & 4523.6\,$|$\,149.4 & 39.34 \\
  Gemini 3.5 Flash     & 4755.2\,$|$\,74.9 & 4853.3\,$|$\,44.4 & 58.20 \\
  Gemini 3.1 Pro       & \textbf{4986.5\,$|$\,4.0} & \textbf{4999.2\,$|$\,0.2} & 81.15 \\
  Kimi K2.5            & 4456.9\,$|$\,171.5 & 4347.4\,$|$\,208.6 & 43.44 \\
  Kimi K2.6            & 4670.2\,$|$\,101.8 & 4541.3\,$|$\,143.5 & 57.38 \\
  Qwen3.5-397B-A17B    & 3332.2\,$|$\,605.3 & 3588.3\,$|$\,494.9 & 31.97 \\
  GPT-5.2              & 4400.8\,$|$\,190.4 & 4324.0\,$|$\,216.7 & 77.05 \\
  Grok 4.3             & 4230.7\,$|$\,249.2 & 4419.3\,$|$\,184.2 & 70.49 \\
  Qwen3.5-122B-A10B    & 3059.8\,$|$\,732.5 & 3587.7\,$|$\,495.1 & 83.61 \\
  Qwen3.5-35B-A3B      & 2120.6\,$|$\,1279.5 & 2084.7\,$|$\,1304.9 & 84.43 \\
  \bottomrule
  \end{tabular*}
\end{table*}

\begin{table*}[ht]
\centering
\small
\caption{\textbf{Taxonomy of observed failure modes.} Each mode represents a distinct category of error exhibited by visual agents during geospatial reasoning tasks.}
\label{tab:failure-modes}
\begin{tabularx}{\textwidth}{@{}l l >{\raggedright\arraybackslash}X l@{}}
\toprule
\textbf{ID} & \textbf{Failure Mode} & \textbf{Description} & \textbf{Affected Stage} \\
\midrule
FM 1 & Vision Failure
     & Errors in interpreting visual input or applying pretrained knowledge, including hallucinated landmark recognition, target object misclassification, and faulty perception of scene content.
     & Vision \\
\addlinespace
FM 2 & Navigational Reasoning Failure
     & Errors in spatial and navigational reasoning, including poor route planning, imprecise movement execution, and lost spatial awareness across observations.
     & Planning / Execution \\
\addlinespace
FM 3 & Tool-Use Failure
     & Errors in tool interface use, including malformed or invalid calls, hallucinated tool capabilities, and refusal to engage due to misunderstood tool semantics.
     & Tool Use \\
\bottomrule
\end{tabularx}
\end{table*}

\begin{figure*}[htbp]
    \centering
    \scriptsize
    \begin{minipage}[b]{0.192\textwidth}\centering
        \textbf{(a)} Qwen3.5-397B\\[1pt]
        \includegraphics[width=0.85\linewidth]{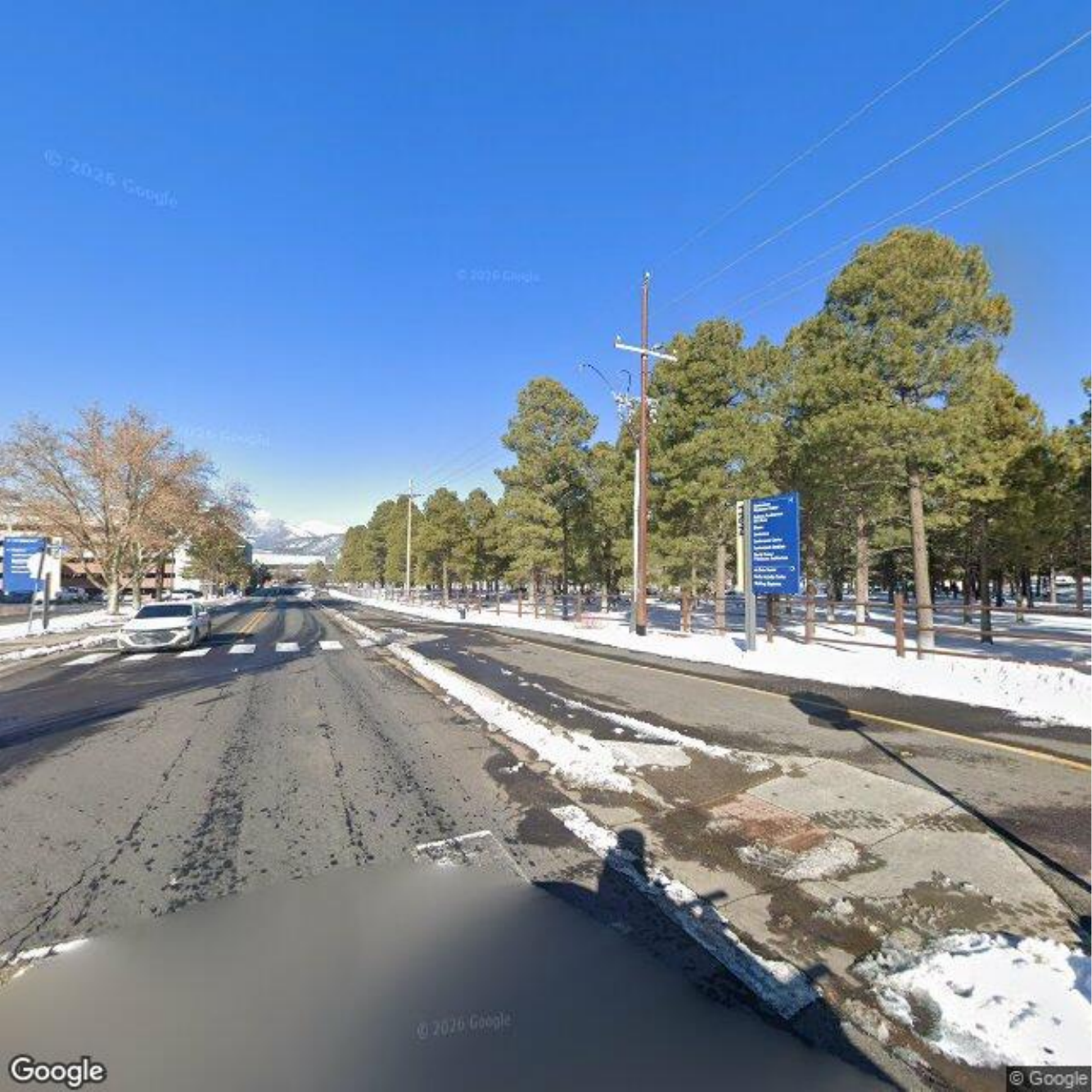}
    \end{minipage}\hfill
    \begin{minipage}[b]{0.192\textwidth}\centering
        \textbf{(b)} GPT-5.2: final view\\[1pt]
        \includegraphics[width=0.85\linewidth]{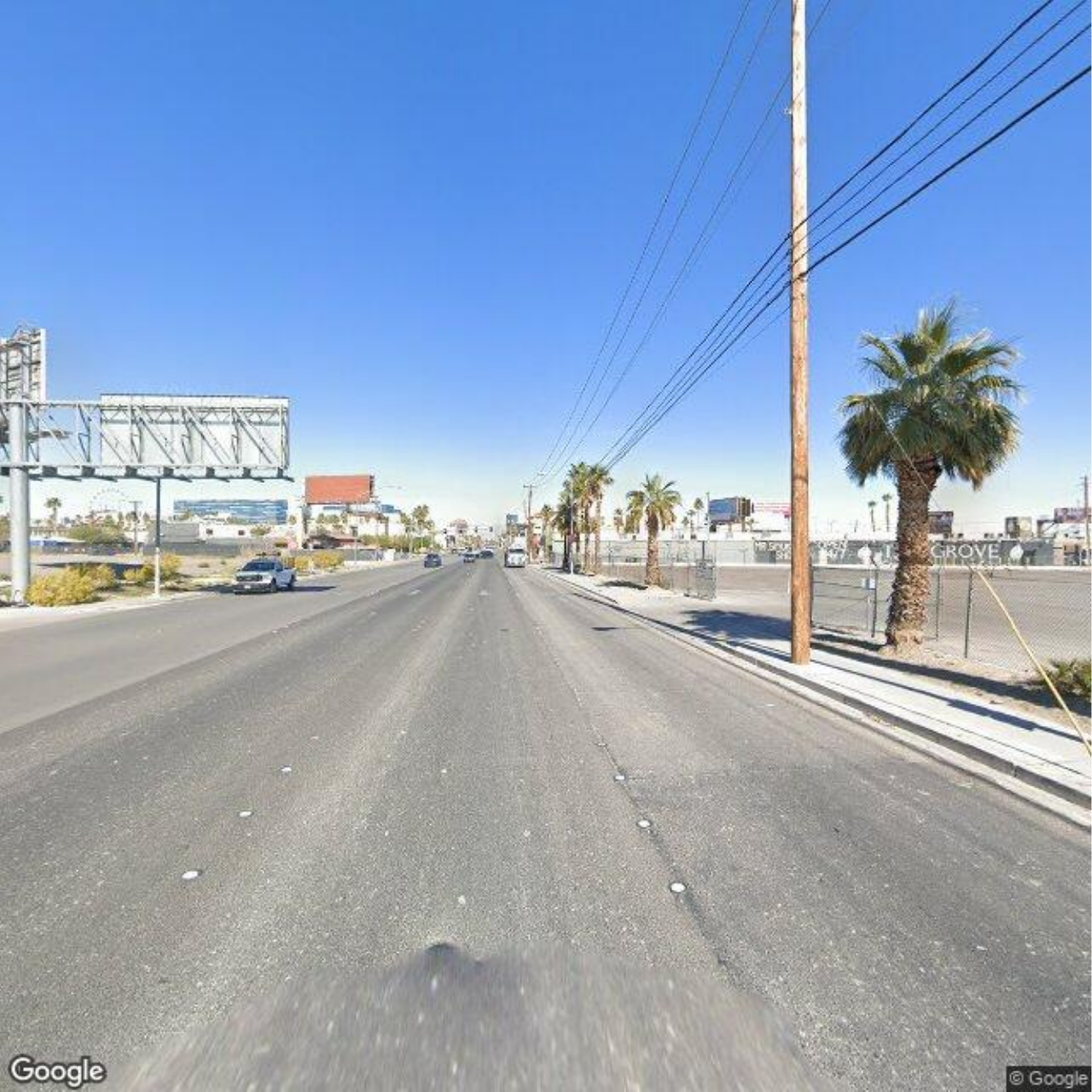}
    \end{minipage}\hspace{2pt}%
    \begin{minipage}[b]{0.192\textwidth}\centering
        \textbf{(b)} intended final view\\[1pt]
        \includegraphics[width=0.85\linewidth]{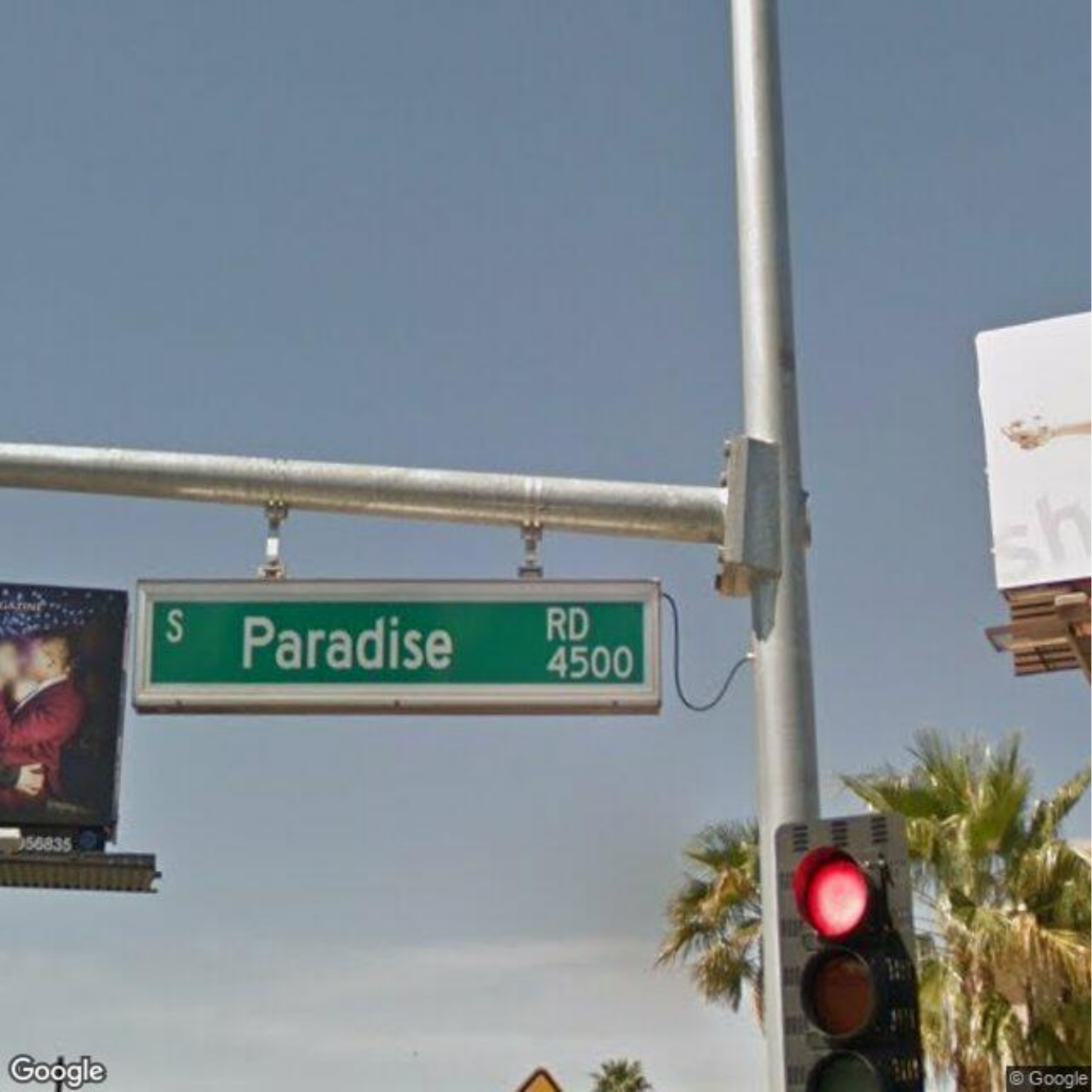}
    \end{minipage}\hfill
    \begin{minipage}[b]{0.192\textwidth}\centering
        \textbf{(c)} Opus 4.6: chosen target\\[1pt]
        \includegraphics[width=0.85\linewidth]{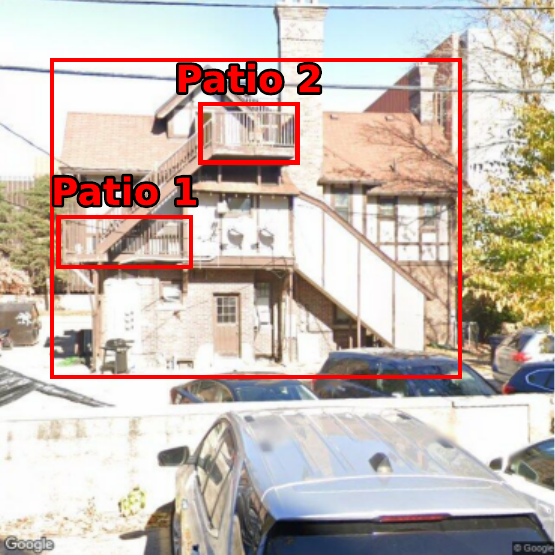}
    \end{minipage}\hspace{2pt}%
    \begin{minipage}[b]{0.192\textwidth}\centering
        \textbf{(c)} intended target\\[1pt]
        \includegraphics[width=0.85\linewidth]{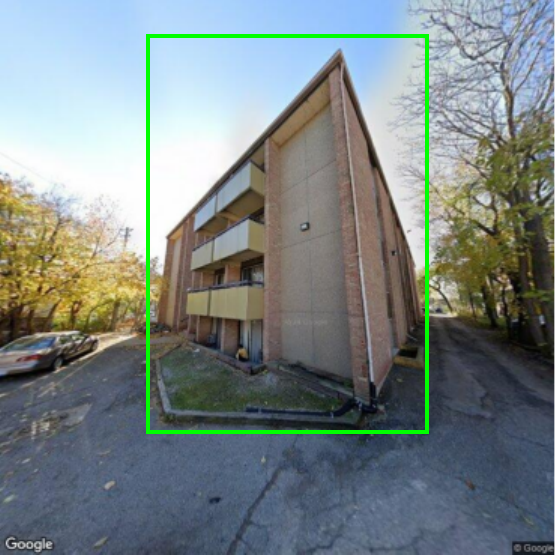}
    \end{minipage}
    \caption{\textbf{Failure-mode examples.} (a)~FM2, \typeTwo{} (``What street are we on?''), intermediate view: the agent overlooks blue signage and crosswalks that suggest a nearby street sign, and answers ``I do not know.'' (b)~FM1, the same question at another location: from its final view, the agent answers ``Las Vegas Boulevard South'' based on the skyline and roadside businesses instead of exploring toward the intended view. (c)~FM1, \typeThree{} (``How many distinct, third floor patios does this building have?''): the model fixates on a house with a gabled roof rather than the intended building. Figures~\ref{fig:fm_4_figure}--\ref{fig:fm_2_figure} give the full prompts and responses.}
    \label{fig:fm_examples}
\end{figure*}

\begin{figure}[htbp]
    \centering
    \setlength{\parskip}{0pt}
    \setlength{\topsep}{0pt}
    \setlength{\partopsep}{0pt}
    \setlength{\itemsep}{0pt}

    \begin{tcolorbox}[fmMode]
        \centering
        \includegraphics[width=\linewidth]{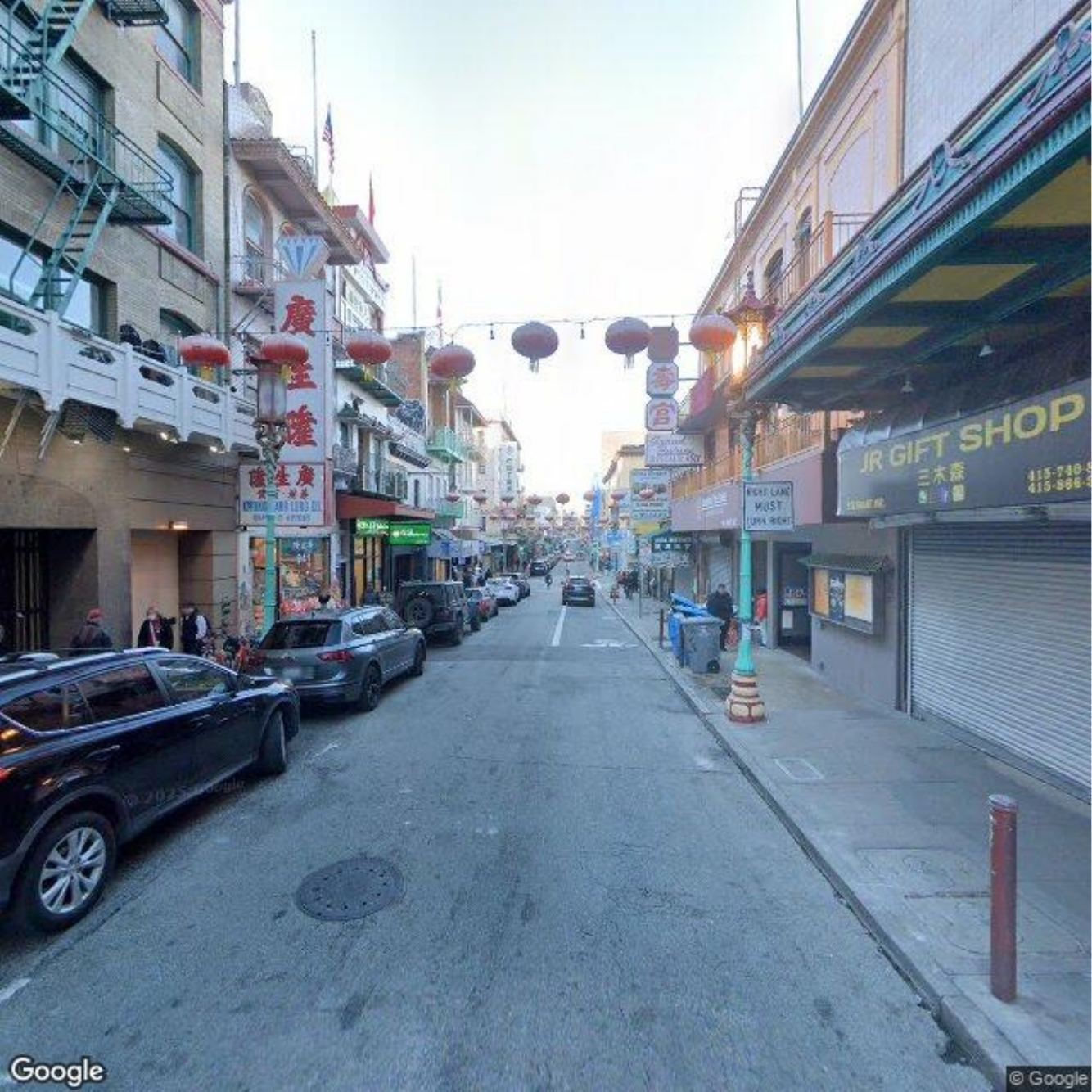}

        \begin{flushleft}
            \textbf{\typeThree Prompt: }How many red lanterns are hanging along the next block south on Grant Avenue?
        \end{flushleft}

        \begin{lstlisting}
'Navigation': move_south x6
'Response': 6
'Reasoning': 'I moved south along Grant Avenue and captured the view of the next block. I counted 6 red lanterns hanging directly above the street on a wire/string, which are the street lanterns and not part of any store decorations.'
\end{lstlisting}
    \end{tcolorbox}

    \caption{\textbf{FM2 (navigational reasoning failure) example: lack of object permanence.} Final view of Qwen3.5-122B-A10B for a \typeThree{} prompt. The model miscounts the lanterns hanging just above and ignores the lanterns it passed on the way to this viewpoint.}
    \label{fig:fm_3_figure}

\end{figure}

\begin{figure}[htbp]
    \centering
    \setlength{\parskip}{0pt}
    \setlength{\topsep}{0pt}
    \setlength{\partopsep}{0pt}
    \setlength{\itemsep}{0pt}

    \begin{tcolorbox}[fmMode]
        \includegraphics[width=\linewidth]{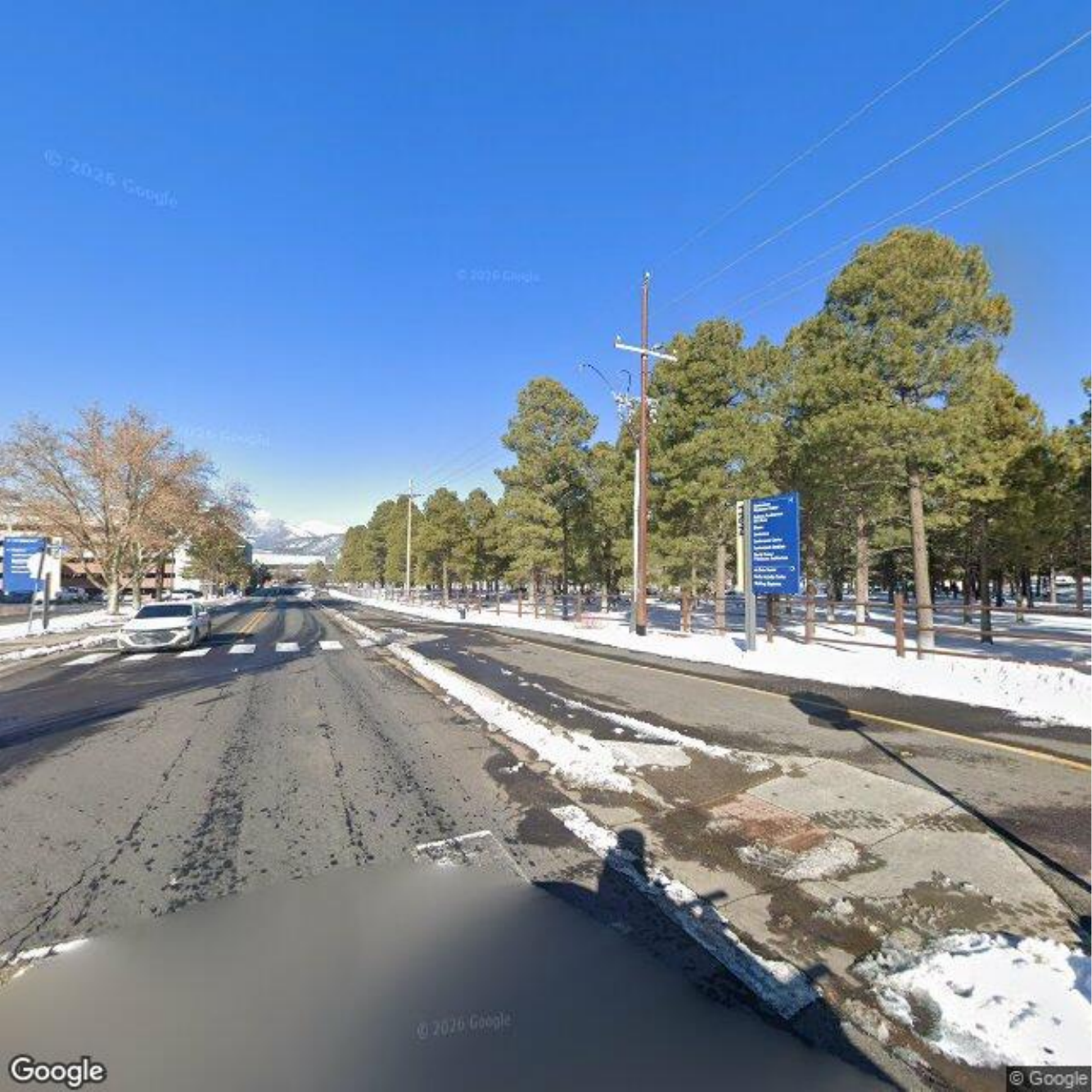}

        \begin{flushleft}
            \textbf{\typeTwo Prompt: }What street are we on?
        \end{flushleft}

        \begin{lstlisting}
'Response': I do not know
'Reasoning': 'I have navigated through the area...but I cannot see any street signs or clear identifiers that would reveal the street name. The images show a road with double yellow lines, snow-covered ground, pine trees on one side, and buildings on the other, with mountains in the distance, but no visible street name signage.'
\end{lstlisting}
    \end{tcolorbox}

    \caption{\textbf{FM2 (navigational reasoning failure) example: poor route planning.} Intermediate view of Qwen3.5-397B-A17B for a \typeTwo{} prompt. The agent overlooks the blue signage and crosswalks, which suggest that a street sign is nearby.}
    \label{fig:fm_4_figure}

\end{figure}

\begin{figure}[htbp]
    \centering
    \setlength{\parskip}{0pt}
    \setlength{\topsep}{0pt}
    \setlength{\partopsep}{0pt}
    \setlength{\itemsep}{0pt}

    \begin{tcolorbox}[fmMode]
        \begin{minipage}{0.49\linewidth}
            \centering
            \textbf{Model Final View}
            \includegraphics[width=\linewidth]{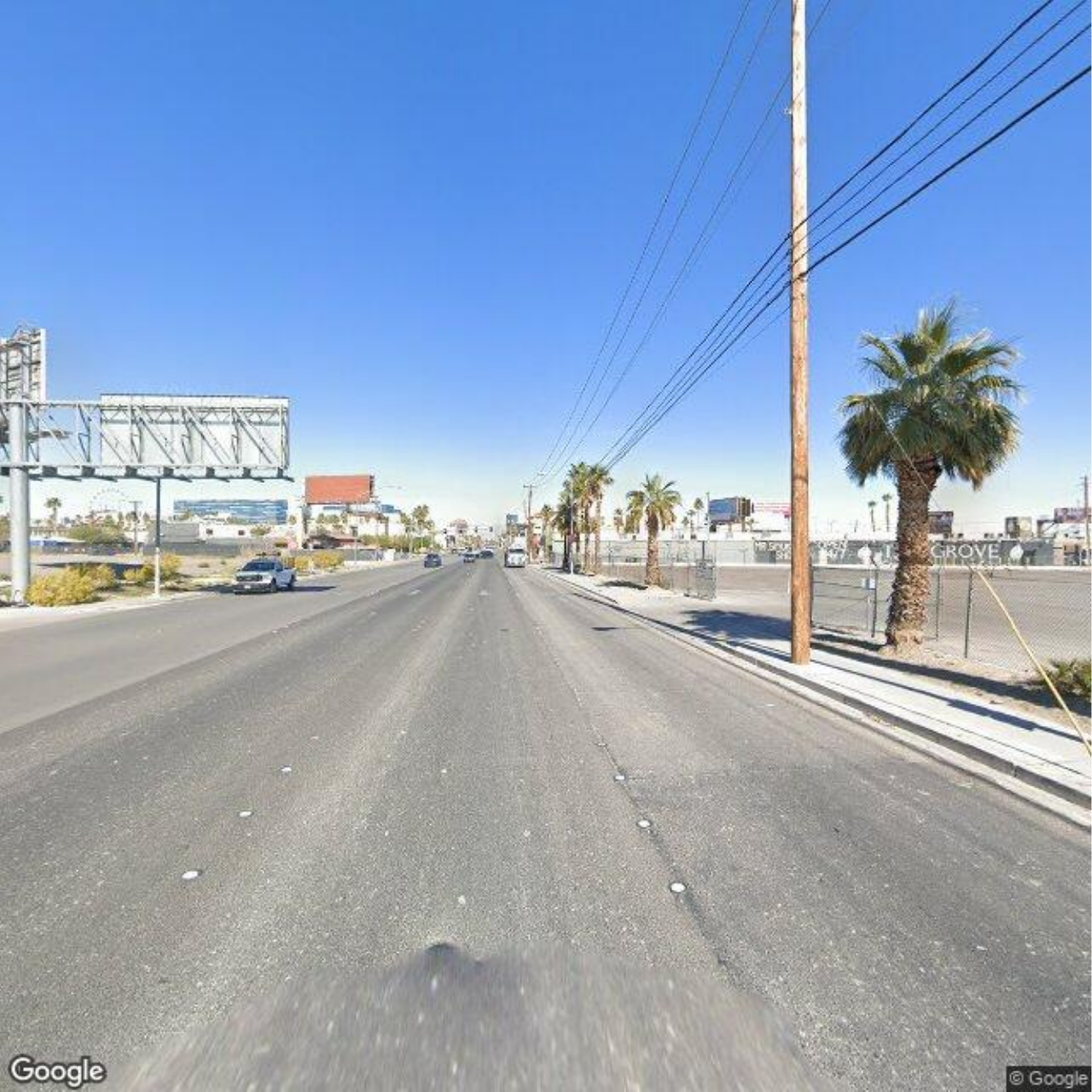}
        \end{minipage}
        \hfill
        \begin{minipage}{0.49\linewidth}
            \centering
            \textbf{Intended Final View}
            \includegraphics[width=\linewidth]{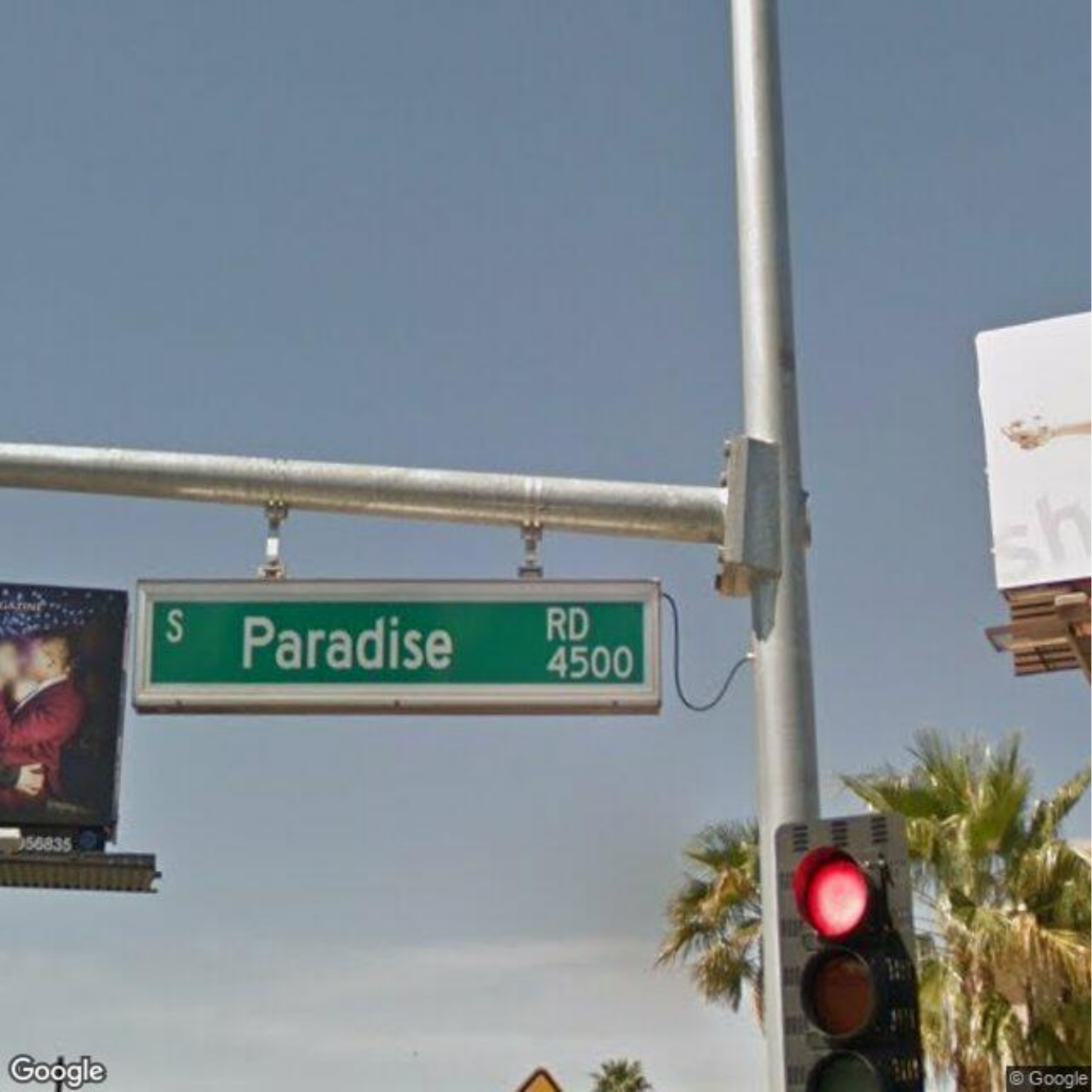}
        \end{minipage}

        \begin{flushleft}
            \textbf{\typeTwo Prompt: }What street are we on?
        \end{flushleft}

        \begin{lstlisting}
'Response': Las Vegas Boulevard South
'Reasoning': 'The skyline and roadside businesses indicate the Las Vegas Strip area, and the corridor shown matches Las Vegas Boulevard South.'
\end{lstlisting}
    \end{tcolorbox}

    \caption{\textbf{FM1 (vision failure) example: hallucinated landmark recognition.} Final view of GPT-5.2 and the intended final view for a \typeTwo{} prompt. Instead of exploring further and grounding its answer in concrete features of the scene, the agent infers its location from sparse evidence and pretrained knowledge.}
    \label{fig:fm_1_figure}

\end{figure}

\begin{figure}[htbp]
    \centering
    \setlength{\parskip}{0pt}
    \setlength{\topsep}{0pt}
    \setlength{\partopsep}{0pt}
    \setlength{\itemsep}{0pt}

    \begin{tcolorbox}[fmMode]
        \begin{minipage}{0.49\linewidth}
            \centering
            \textbf{Intended Target}
            \includegraphics[width=\linewidth]{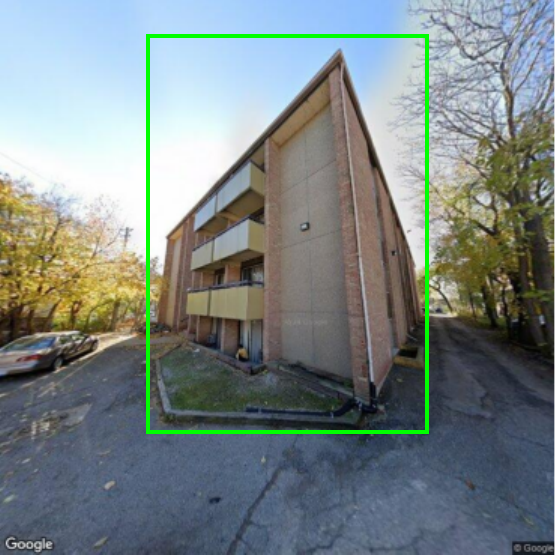}
        \end{minipage}
        \hfill
        \begin{minipage}{0.49\linewidth}
            \centering
            \textbf{Model Target}
            \includegraphics[width=\linewidth]{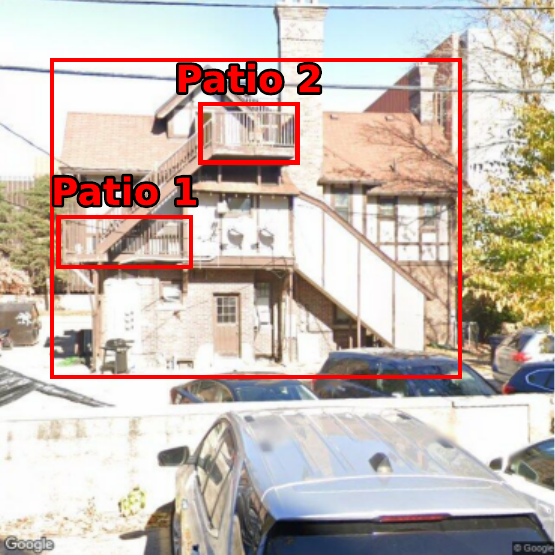}
        \end{minipage}
        \vspace{1pt}
        \begin{flushleft}
            \textbf{\typeThree Prompt: }You are in the back parking lot of an apartment building. How many distinct, third floor patios does this building have?
        \end{flushleft}

    \end{tcolorbox}

    \caption{\textbf{FM1 (vision failure) example: fixation on a distractor.} Intended target and the target chosen by Claude Opus 4.6 for a \typeThree{} prompt. The model fixates on the house with a gabled roof.}
    \label{fig:fm_2_figure}

\end{figure}

\begin{figure}[htbp]
    \centering
    \setlength{\parskip}{0pt}
    \setlength{\topsep}{0pt}
    \setlength{\partopsep}{0pt}
    \setlength{\itemsep}{0pt}

    \begin{tcolorbox}[fmMode]
        \begin{flushleft}
            \textbf{\typeOne Prompt: }Identify the most precise coordinates of this starting location.
        \end{flushleft}

        \begin{lstlisting}
'Response': I do not know
'Reasoning': '...The available tools only allow me to navigate and capture street view images, but I cannot analyze visual content to identify geographic coordinates.'
\end{lstlisting}
    \end{tcolorbox}

    \caption{\textbf{FM3 (tool-use failure) example.} Response of Qwen3.5-122B-A10B to a \typeOne{} prompt before taking any action. The agent misunderstands what \texttt{capture\_view} does (Section~\ref{sec:eval_setup}).}
    \label{fig:fm_5B_figure}

\end{figure}

\end{document}